\documentclass[a4paper,fleqn]{cas-dc}

\usepackage[authoryear]{natbib}
\usepackage{booktabs}
\usepackage{multirow}
\usepackage{amsmath}
\usepackage{amssymb}
\usepackage{graphicx}
\usepackage{xcolor}
\usepackage[most]{tcolorbox}
\usepackage{simpleicons}
\usepackage{hyperref}

\newtcolorbox{rqbox}[1]{
  enhanced, breakable,
  colback=teal!3, colframe=teal!60!black,
  boxrule=0.5pt, arc=1pt,
  left=5pt, right=5pt, top=3pt, bottom=3pt,
  fonttitle=\bfseries\small, coltitle=white,
  title=#1,
  fontupper=\small,
}

\begin{document}
\let\WriteBookmarks\relax
\def\floatpagepagefraction{1}
\def\textpagefraction{.001}

\shorttitle{Fine-Grained Emotion Classification from App Reviews}

\shortauthors{Motger et al.}

\title [mode = title]{Fine-Grained Emotion Classification from Mobile App Reviews: An Empirical Study with Large Language Models}

\tnotemark[1]
\tnotetext[1]{Replication package including source code, evaluation results, datasets, prompts and fine-tuned models is released with this article.}

\author[1]{Quim Motger}
\cormark[1]
\ead{joaquim.motger@upc.edu}
\credit{Conceptualization, Methodology, Software, Investigation, Data curation, Validation, Writing -- original draft}

\author[1]{Carlota Catot}
\ead{carlota.catot@upc.edu}
\credit{Software, Investigation, Data curation, Validation, Writing -- review \& editing}

\author[1]{Marc Oriol}
\ead{marc.oriol@upc.edu}
\credit{Conceptualization, Methodology, Validation, Writing -- review \& editing}

\affiliation[1]{organization={Department of Service and Information System Engineering, Universitat Polit\`ecnica de Catalunya},
            country={Spain}}

\cortext[1]{Corresponding author}

\begin{abstract}
\textbf{Context:}
Fine-grained emotion classification of mobile app reviews enables requirements engineering activities that go beyond polarity-based opinion mining, including emotionally informed issue prioritisation and feature-oriented feedback analysis. However, automatic fine-grained emotion extraction from app reviews remains understudied.
\textbf{Objectives:}
Building on a previously published annotation framework and human-labelled ground truth adapted from Plutchik's taxonomy, this paper investigates how large language models can be leveraged for automatic multi-label emotion classification under severe class imbalance.
\textbf{Methods:}
We compare encoder-only fine-tuning under multi-label and binary-ensemble formulations, decoder-only zero- and few-shot prompting across open-source and proprietary models, and a catalogue of imbalance mitigation strategies (loss reweighting, resampling, generative data augmentation), with the synthetic-review generator and prompting strategy selected via an intrinsic augmentation-utility ranking.
\textbf{Results:}
Fine-tuned encoders trail the best decoder-only few-shot prompting (macro-F1 $0.642$) by a wide margin at baseline (multi-label: $0.387$; binary ensemble: $0.450$); pairing the best multi-label encoder with generative data augmentation and positive-weighted loss closes most of this gap ($+0.204$) at up to three orders of magnitude lower inference latency than the decoders, with the largest gains on the rarest emotions, from undetected to gains of up to $+0.501$ F1. 
\textbf{Conclusion:}
Large language models make fine-grained, multi-label emotion classification of app reviews feasible for requirements engineering pipelines, with modest macro-F1, and the best formulation and mitigation strategy are backbone- and formulation-dependent. We release the experimental pipeline, synthetic corpora, and fine-tuned checkpoints for replication and reuse.
\end{abstract}


\begin{keywords}
mobile app reviews \sep emotion classification \sep large language models \sep class imbalance \sep data augmentation \sep requirements engineering
\end{keywords}

\maketitle

\section{Introduction}\label{sec:intro}

Mining mobile app reviews has become a consolidated source of user feedback for requirements engineering (RE) and software evolution~\citep{Dabrowski2022,Pagano2013,Maalej2015}.
Prior work extracts ratings~\citep{Pagano2013}, review types such as bug reports and feature requests~\citep{Maalej2015}, topics and aspects~\citep{Alturaief2021,Alturayeif2023}, app features~\citep{Guzman2014,Johann2017,Motger2024TFREX,Motger2025TFREXEMSE}, and sentiment polarity~\citep{Guzman2014,Araujo2022,Lin2022}.
Polarity remains the dominant affective descriptor in app-store mining~\citep{Lin2022,Zhang2025Sentiment}, yet positive/negative/neutral labels compress distinct user states that matter for RE decision-making.
Some examples include excitement about a feature, disappointment caused by a defect, explicit rejection of an app, or anticipation of a missing capability~\citep{Motger2025Emotions,Cheng2023,Hidellaarachchi2023}.

Fine-grained emotion analysis has been studied in other domains, such as social media and psychotherapy transcripts~\citep{AMEER2023118534,Lalk2025}, and more broadly in software-related opinion mining~\citep{Lin2022}, but automatic multi-label emotion classification of English mobile app reviews remains scarce~\citep{Motger2025Emotions}.
Existing emotion pipelines for app feedback are often lexicon-based, proprietary, or tied to unpublished datasets~\citep{Motger2025Emotions}, which limits comparative evaluation and industrial reuse.
Large language models (LLMs) based on the Transformer architecture~\citep{Vaswani2017,Zhao2023,Hou2024} offer two complementary routes for this task: (i)~encoder-only models fine-tuned as supervised sequence classifiers~\citep{Devlin2019,Liu2019RoBERTa,Motger2024TFREX}, and (ii)~decoder-only generative models used through zero- and few-shot prompting~\citep{Brown2020,Zhang2025Sentiment}.
In addition to the lack of datasets, both families confront a severe multi-label class imbalance, with frequent emotions such as Joy or \emph{Sadness} coexisting with rare ones such as \emph{Fear} or \emph{Anger}~\citep{Motger2025Emotions,Cui2025Weighted}.

In prior work, we adapted Plutchik's eight basic emotions~\citep{Plutchik2001} to the mobile app review domain, produced annotation guidelines, released a human-labelled ground truth of 1{,}090 sentences (excluding rejected ones) from 257 apps spanning 10 Google Play categories, and characterised multiple emotion annotation challenges~\citep{Motger2025Emotions}.
That study explicitly left open the design of automatic extraction pipelines, recommending multi-label classifiers or binary ensembles, encoder- and decoder-only settings, and imbalance mitigation including synthetic generation~\citep{Motger2025Emotions}.
This paper addresses that gap.

Our contributions are:
\begin{itemize}
\item An empirical comparison of encoder-only fine-tuning under multi-label and binary-ensemble formulations, using multiple encoder backbones. 
\item An empirical comparison of decoder-only zero- and few-shot emotion classification across a curated set of open-source and proprietary generative models.
\item A systematic evaluation of imbalance mitigation strategies (i.e., class-weighted losses, focal and adaptive focal losses, undersampling, and generative data augmentation) on minority and majority emotions. 
\item An augmentation-utility analysis that ranks generator--prompt configurations by intra-set diversity, novelty, and emotion fidelity before synthetic reviews are used for fine-tuning.
\item An augmented dataset of 3{,}308 labelled texts (1{,}090 human-annotated review sentences from previous work and 2{,}218 synthetic reviews generated in this study), rebalancing under-represented emotion labels.
\item A replication package and Hugging Face release of fine-tuned models.
\end{itemize}

Section~\ref{sec:background} summarises background, Section~\ref{sec:design} presents the research design, and Section~\ref{sec:system} the system design.
Section~\ref{sec:results} reports the results, Section~\ref{sec:discussion} discusses findings and threats to validity, Section~\ref{sec:rel-work} positions related work, and Section~\ref{sec:conclusions} highlights main conclusions.

\section{Background}\label{sec:background}

\subsection{Opinion Mining in Software Engineering}\label{sec:op-min-b}

Opinion mining (often used interchangeably with sentiment analysis) denotes computational methods that recover subjective information from text, ranging from polarity and emotion to politeness, toxicity and related affective constructs~\citep{Lin2022}.
In software engineering (SE), such methods have been applied to heterogeneous artefacts, including issue discussions, commit messages, pull-request conversations and end-user feedback~\citep{Lin2022,Zhang2025Sentiment}.
Their value for RE lies in turning large volumes of informal language into structured signals that support elicitation, prioritisation, validation and release planning~\citep{Hidellaarachchi2023,Cheng2023,Dabrowski2022}.
In this context, app-store reviews are a particularly rich, continuously growing source, as users spontaneously report defects, request features, praise quality attributes and express dissatisfaction at scale~\citep{Pagano2013,Maalej2015}.

Within app-review mining, sentiment analysis remains dominated by polarity, sometimes combined with aspect or feature mentions~\citep{Guzman2014,Araujo2022,Lin2022,Zhang2025Sentiment}.
Polarity is useful but lossy: reviews labelled ``positive'' may encode excitement about a feature, reliance on the app for personal goals, or generic appraisal, while ``negative'' reviews may encode disappointment about defects, rejection of the product, or hostility~\citep{Motger2025Emotions}.
Fine-grained emotion taxonomies, notably Plutchik's eight primary emotions~\citep{Plutchik2001}, provide a vocabulary that better matches these distinctions, and this study addresses automatic multi-label classification over that vocabulary.

\subsection{Emotion-Annotated App Review Datasets}\label{sec:emo-datasets-b}

Despite growing interest in affective signals from app stores, publicly reusable corpora that combine \emph{mobile app reviews} with \emph{human emotion labels} remain scarce~\citep{Motger2025Emotions,Lin2022}.
Existing efforts differ along dimensions that matter for supervised and prompted classification: language, label inventory and theoretical grounding, annotation unit (sentence vs.\ full review), whether multiple emotions may co-occur, how labels were produced (human guidelines vs.\ lexicons or proprietary services), documentation of the annotation process, and availability for replication.
Table~\ref{tab:emo-datasets} summarises representative resources along these criteria, except label co-occurrence, which we discuss below.

\begin{table*}[t]
\centering
\caption{Comparison of emotion-oriented resources for (mobile) app or closely related software reviews.
``Avail.'' indicates whether a reusable emotion-labelled corpus is publicly released.}
\label{tab:emo-datasets}
\small
\begin{tabular}{@{}p{2.8cm}p{1cm}p{1.7cm}p{1.55cm}p{1.55cm}p{2.3cm}p{2.4cm}c@{}}
\toprule
\textbf{Resource} & \textbf{Lang.} & \textbf{Labels} & \textbf{Unit} & \textbf{Size} & \textbf{Annotation} & \textbf{Guidelines} & \textbf{Avail.} \\
\midrule
\citet{Motger2025Emotions}
  & EN
  & Plutchik (8) + \emph{Neutral}
  & Sentence & 1{,}090
  & Human ($n=5$)
  & Publicly released guidelines + examples
  & Yes \\
\citet{Riccosan2023}
  & ID
  & Shaver (5) + \emph{Neutral}
  & Sentence & 21{,}697
  & Human ($n=2$)
  & Paper-level description
  & Yes \\
\citet{Siqueira2024}
  & PT
  & Ekman (6) + \emph{Neutral}
  & Review & 3{,}000
  & Human ($n=3$)
  & Paper-level description
  & Yes \\
\citet{Singh2022}
  & EN
  & Plutchik
  & Review & 2{,}000
  & Automatic (lexicon-based)
  & Tool-oriented
  & No$^\ddagger$ \\
\citet{Keertipati2016}
  & EN
  & LIWC negative emotions
  & Review & 4{,}442
  & Automatic (lexicon-based)
  & Tool-oriented
  & No$^\ddagger$ \\
\citet{Malgaonkar2019}
  & EN
  & Ekman (6)
  & Review & 52{,}705
  & Automatic (lexicon-based)
  & Tool-oriented
  & No$^\ddagger$ \\
\citet{Savarimuthu2023}
  & EN
  & Custom
  & Review & 35{,}576
  & Automatic (machine learning-based)
  & Tool-oriented
  & No$^\ddagger$ \\
\citet{Cabellos2022}
  & ES
  & Liew \& Turtle (28)
  & Review & 1{,}560
  & Human ($n=2$)
  & Paper-level description
  & No \\
\bottomrule
\multicolumn{8}{@{}l@{}}{\footnotesize $^\ddagger$Emotion labels used as study features; no public gold emotion corpus for classifier training.}
\end{tabular}
\end{table*}

Two public corpora are closest in intent to ours but differ critically for English RE-oriented multi-label classification.
\citet{Riccosan2023} release a large Indonesian sentence-level collection ($\approx$21.7k) with five emotion classes (anger, fear, sadness, happiness, love) plus neutral, grounded in Shaver's basic-emotion hierarchy, plus sentiment polarity.
It is not English, omits Plutchik categories that are salient in app feedback such as \emph{Trust} and \emph{Anticipation}~\citep{Motger2025Emotions}, assigns a single emotion per sentence, and documents annotation mainly through pairwise agreement ($\kappa{\approx}0.61$) without rich domain guidelines or disagreement-resolution analysis.
\citet{Siqueira2024} provide 3{,}000 Brazilian Portuguese Play Store reviews labelled with an Ekman-style inventory plus \emph{Neutral} and polarity, using three annotators who reconcile disagreements by consensus.
Relative to our setting, the language differs, labels are assigned at review granularity instead of sentence, emotions are mutually exclusive, \emph{Trust}/\emph{Anticipation} are again absent, and no reusable guideline artefact is released.

Beyond these, most English studies that mention emotions in app reviews rely on proprietary or non-reusable data, or on automatic lexicons and APIs instead of gold multi-label annotations by experts.
\citet{Singh2022} analyse $\approx$2k English mobile-learning reviews with a Plutchik-aligned, lexicon-assisted process, but the annotated corpus is not publicly available for classifier benchmarking.
\citet{Keertipati2016} and \citet{Malgaonkar2019} derive affective features with LIWC and WordNet-based lexicons, respectively.
\citet{Savarimuthu2023} obtain emotion descriptors from IBM Watson's Tone Analyzer for a data-waste prediction task.
Such pipelines demonstrate the usefulness of emotion signals for SE analytics, yet they do not yield an open, guideline-grounded gold standard for supervised multi-label evaluation.
\citet{Cabellos2022} manually interpret Spanish Steam reviews of a video game under Liew and Turtle's 28-emotion taxonomy, but target a different product genre and release no classification-ready dataset.

Against this landscape, we select the ground truth of~\citet{Motger2025Emotions} as the primary experimental dataset for the following reasons.
(1)~\emph{Language and RE scope:} English reviews across 10 Google Play categories, aligned with the bulk of app-store mining research~\citep{Dabrowski2022}.
(2)~\emph{Emotion scope:} Plutchik's eight primary emotions plus \emph{Neutral}, including \emph{Trust} and \emph{Anticipation}, which prior annotation work argued are particularly informative for feature reliance and unmet expectations in app reviews~\citep{Motger2025Emotions,Plutchik2001}.
(3)~\emph{Task formulation:} sentence-level, multi-label annotation matching co-occurring affective states in user feedback~\citep{Tsoumakas2007}.
(4)~\emph{Annotation quality:} an iterative human process with five annotators, publicly released domain-adapted guidelines (definitions, disambiguation rules and illustrative examples), and documented challenges and agreement analyses. 
(5)~\emph{Openness:} labelled sentences and guidelines are openly shared, enabling replication.

\subsection{Sequence Classification with LLMs}\label{sec:seq-class-b}

Modern sequence classification is largely built on the Transformer architecture~\citep{Vaswani2017}.
Encoder-only models such as BERT, RoBERTa, DistilBERT, DeBERTa and XLNet~\citep{Devlin2019,Liu2019RoBERTa,Sanh2019DistilBERT,He2023DeBERTaV3,Yang2019XLNet} produce bidirectional contextual representations and are typically adapted to downstream tasks by attaching a lightweight classification head and fine-tuning on labelled data~\citep{Hou2024}.
This paradigm has become a default for SE text classification and information extraction, including app-review analysis~\citep{Hou2024,Motger2024TFREX,Motger2025TFREXEMSE}.
When several labels may apply simultaneously, the task is multi-label rather than mutually exclusive multi-class~\citep{Tsoumakas2007}.
Two standard formulations are used: a single multi-output classifier (e.g., independent sigmoid activations trained with binary cross-entropy) and problem transformation into an ensemble of binary one-vs-rest classifiers~\citep{Tsoumakas2007}.
The former shares representation learning across labels; the latter can specialise decision boundaries per emotion but multiplies training and inference cost.

Decoder-only (generative) LLMs offer an alternative adaptation path based on prompt engineering~\citep{Brown2020,Zhao2023}.
In zero-shot settings, the model receives task instructions and label definitions; in few-shot settings, a small number of labelled exemplars is added in context~\citep{Brown2020}.
Recent SE studies on sentiment-related tasks report that prompted generative models can approach specialised systems, yet remain sensitive to prompt wording, model family, decoding parameters and the subjectivity of affective labels~\citep{Zhang2025Sentiment,Hou2024}.
Beyond classification, the same generative models can synthesise additional labelled text, which is increasingly explored as a data-centric complement when annotated corpora are small or skewed~\citep{AnabyTavor2020,Motger2025Emotions}.

\subsection{Data Imbalance Mitigation Strategies}\label{sec:dat-imb-b}

Class imbalance arises when label frequencies differ substantially, so that learning is dominated by majority classes and minority classes are under-detected~\citep{He2009Imbalance}.
In multi-label emotion data, a single sentence may mix frequent and rare emotions, so naive undersampling of ``majority instances'' can simultaneously remove minority emotions~\citep{Charte2015MLSMOTE,Cui2025Weighted}.
App-review emotion corpora exhibit precisely this profile~\citep{Motger2025Emotions}.

Mitigation strategies are commonly grouped into algorithm-level and data-level families~\citep{He2009Imbalance}.
Algorithm-level approaches keep the training sample unchanged and reweight the learning objective: class-dependent weights in binary cross-entropy (BCE) upweight rare positives~\citep{Cui2025Weighted}, while focal loss further down-weights easy examples so that training focuses on hard or infrequent cases~\citep{Lin2017Focal,SemEval2025Focal}.
Adaptive variants adjust focusing behaviour from label frequencies to reduce manual tuning~\citep{Labd2025AFL}.
Data-level approaches modify the training distribution through undersampling of majority cases~\citep{He2009Imbalance} or synthetic oversampling such as the multi-label MLSMOTE~\citep{Charte2015MLSMOTE}.
A complementary data-level trend is generative augmentation: large language models produce additional minority-class texts that can be injected into fine-tuning, provided the synthetic samples remain diverse, non-redundant and faithful to the target label~\citep{AnabyTavor2020,Motger2025Emotions}.
Empirical evidence on multi-label emotion benchmarks suggests that weighting and focalisation can help, but improvements on the rarest classes are often modest and highly configuration-dependent~\citep{Cui2025Weighted,SemEval2025Focal}. 

\section{Research Design}\label{sec:design}

\subsection{Goal and Research Questions}\label{sec:rq}

The goal of this research is to \textbf{analyse} and \textbf{compare} how large language models can be leveraged for automatic fine-grained emotion classification from mobile app reviews, with particular attention to multi-label formulation and data imbalance, \textbf{from the point of view of} software engineering teams and researchers, \textbf{in the context of} RE-oriented opinion mining of app-store feedback.

Following evaluation-oriented research questions for software engineering~\citep{Shaw2003}, we formulate:

\begin{description}
\item[\textbf{RQ1.}] How effective are encoder-only LLMs at multi-label emotion classification of mobile app reviews under multi-label and binary-ensemble formulations?
\item[\textbf{RQ2.}] How effective are decoder-only LLMs at zero-shot and few-shot emotion classification of mobile app reviews?
\item[\textbf{RQ3.}] How can generative LLMs be selected and configured to produce synthetic reviews for under-represented emotions that are useful for fine-tuning?
\item[\textbf{RQ4.}] Which data-imbalance mitigation strategies most improve encoder-only emotion classification performance?
\end{description}


To address RQ1, we fine-tune a catalogue of encoder-only models as (A)~a single multi-label classifier and (B)~an ensemble of binary classifiers, using the human ground truth from~\citet{Motger2025Emotions}, and report per-emotion and aggregate correctness together with a cost/latency profile.

To address RQ2, we evaluate open-source and proprietary decoder-only models as zero-shot and few-shot classifiers instructed with the emotion guidelines from~\citet{Motger2025Emotions}, assessing sensitivity to decoding temperature alongside the prompting regime, with metrics and cost profile comparable to RQ1.

To address RQ3, we generate synthetic reviews for every under-represented emotion label under multiple generators and prompting strategies, rank configurations with embedding-based augmentation-utility metrics, and feed the selected pipeline into RQ4 at controlled injection sizes. 

To address RQ4, we instantiate imbalance mitigation on the best encoder settings from RQ1, covering loss reweighting, focal variants, undersampling, and generative augmentation using the RQ3 pipeline.

\subsection{Research Method}\label{sec:method}

\paragraph{Evaluation protocol.}
All supervised experiments use 10-fold multi-label stratified cross-validation~\citep{Sechidis2011}, with the same random seed across configurations, over the 938 sentences that remain after setting aside a fixed stratified tuning holdout of 152 sentences (Section~\ref{sec:encoder-class}).
Resampling and synthetic augmentation are applied inside each training fold only, never on the held-out fold, to avoid leakage. 
Decoder-only classification (RQ2) needs no tuning holdout and is evaluated against the same evaluation set. 

\paragraph{Scope of analysis.}
RQ1 and RQ2 compare two deployment paradigms.
Encoder-only systems are specialised classifiers, requiring labelled data and gradient-based fine-tuning, but yield compact checkpoints with low inference cost. This setting is aligned with continuous mining of large app-review streams~\citep{Motger2024TFREX,Hou2024}.
Decoder-only systems are evaluated in the prompting setting that practitioners typically adopt when labelled fine-tuning of large generative models is impractical (API-only access, cost, or lack of local GPU capacity).
They perform zero- and few-shot classification guided by the same annotation guidelines used by human annotators~\citep{Brown2020,Zhang2025Sentiment,Motger2025Emotions}, and no decoder weights are updated.
Consequently, absolute metric gaps between RQ1 and RQ2 should be interpreted as trade-offs among labelling effort, computational cost, model size and operational constraints. 
Both paradigms share the same label space, fold protocol, metrics and guideline source document.

\paragraph{Labels.}
We classify the eight Plutchik emotions~\citep{Plutchik2001} (see Section~\ref{sec:encoder-class}) plus \emph{Neutral}, as defined in the annotation guidelines~\citep{Motger2025Emotions}.
The Reject label from the original schema (i.e., sentences which could not be processed by the original experts) is dropped.
Emotions are not mutually exclusive: a sentence may carry multiple labels (at most three; Section~\ref{sec:dataset}).

\paragraph{Metrics.}
For each emotion we report Precision, Recall and F1.
As aggregates we report macro-averaged Precision/Recall/F1 and micro-F1.
Macro-F1 is the primary decision metric when comparing imbalance strategies (equal weight to minority emotions); micro-F1 and per-emotion F1 contextualise overall behaviour and class-wise failure modes~\citep{Tsoumakas2007,Cui2025Weighted}.

\paragraph{Experimental factors.}
The experimental framework is designed so that model selection, classification formulation, prompting regime, imbalance method and hyperparameters can be varied independently (Table~\ref{tab:factors}).
The encoder learning rate, which lacks a literature-backed default for this task, is tuned on a fixed holdout excluded from cross-validation; other defaults follow prior work (Section~\ref{sec:system}). Decoding temperature and imbalance-specific parameters (e.g., undersampling cut-offs, synthetic injection size) are swept and reported in full as sensitivity analyses.

\begin{table*}[t]
\centering
\caption{Main experimental factors.}
\label{tab:factors}
\small
\begin{tabular}{@{}llp{10.2cm}@{}}
\toprule
\textbf{Factor} & \textbf{RQ} & \textbf{Levels (summary)} \\
\midrule
Encoder backbone & RQ1, RQ4 & See Table~\ref{tab:models} (base/large variants) \\
Classification head & RQ1, RQ4 & Multi-label (single model) vs.\ binary ensemble \\
Encoder training hyperparameters & RQ1, RQ4 & Learning rate (tuned per backbone); epochs 3; batch size 8; max length 512 \\
Decoder model & RQ2, RQ3 & See Table~\ref{tab:models} (open-source and proprietary) \\
Prompting regime & RQ2, RQ3 & Zero-shot; few-shot (guidelines); few-shot (guidelines + dataset samples) \\
Decoding temperature & RQ2 & Sensitivity grid $\{0.0, 0.3, 0.7\}$ (default $0.0$ for primary tables) \\
Imbalance method & RQ4 & Baseline; class weights; focal; adaptive focal; undersampling; generative augmentation; selected combinations \\
Imbalance hyperparameters & RQ4 & Focal $\gamma{=}2.0$, mean reduction (fixed); undersampling cut-off and synthetic injection size in bins of 50 \\
\bottomrule
\end{tabular}
\end{table*}

\subsection{Dataset}\label{sec:dataset}

We reuse the human-annotated ground truth of~\citet{Motger2025Emotions} without additional filters, for the comparative reasons detailed in Section~\ref{sec:emo-datasets-b}.
The dataset contains 1{,}090 review sentences from 257 mobile apps across 10 Google Play categories, labelled at sentence level following publicly released guidelines (included in the replication package).
Label frequencies are strongly skewed (Table~\ref{tab:rq1-per-emotion}): majority emotions (e.g., \emph{Joy}, \emph{Sadness}) coexist with scarce minorities (e.g., \emph{Fear}, \emph{Anger}, \emph{Disgust}), which is the primary driver of RQ3--RQ4.
Multi-label co-occurrence is present but limited: $85.8\%$ of sentences carry exactly one emotion, $13.8\%$ carry two, and $0.5\%$ carry three, which is both the ground for the maximum cardinality of three used at inference (Section~\ref{sec:encoder-class}) and, as that cardinality is rarely reached, the reason inference uses a confidence threshold. 

\section{System Design}\label{sec:system}

\subsection{Encoder-only Classification}\label{sec:encoder-class}

\paragraph{Task formulation.}
Let $E=\{\text{\emph{Joy}},\allowbreak\text{\emph{Trust}},\allowbreak\text{\emph{Fear}},\allowbreak\text{\emph{Surprise}},\allowbreak\text{\emph{Sadness}},\allowbreak\text{\emph{Disgust}},\allowbreak\text{\emph{Anger}},\allowbreak\text{\emph{Anticipation}},\allowbreak\text{\emph{Neutral}}\}$. 
Given a review sentence $x$, the encoder produces contextual representations and a classification head outputs logits $z\in\mathbb{R}^{|E|}$.
Two classifier formulations (\textit{A}, \textit{B}) are compared. 

\begin{itemize}
    \item \textit{Multi-label classifier (A).} We attach a linear layer with $|E|$ outputs and train with a multi-label objective (Section~\ref{sec:data-imbalance}). At inference, probabilities $p=\sigma(z)$ are obtained via sigmoid. Each label is assigned independently via a decision threshold $p_c \geq \theta = 0.5$; if no label crosses the threshold we keep the single most confident one (never predicting an empty set), and if more than three cross it we keep only the three most confident, matching the maximum cardinality observed in the ground truth (Section~\ref{sec:dataset}).
    \item \textit{Binary ensemble (B).} We train one binary classifier per emotion $c\in E$, using presence/absence of $c$ as the target. At inference, each binary head yields a confidence score; we assemble the multi-label prediction with the same threshold-then-cap rule as (A), so the two formulations differ only in how the per-emotion probabilities are produced (one shared model vs.\ nine independent ones). 
\end{itemize}



\paragraph{Model catalogue.}
Table~\ref{tab:models} lists all LLMs used in this study.
For encoder-only fine-tuning (RQ1, RQ4), we evaluate BERT$_\text{base/large}$~\citep{Devlin2019}, RoBERTa$_\text{base/large}$~\citep{Liu2019RoBERTa}, DistilBERT~\citep{Sanh2019DistilBERT}, DeBERTa-v3$_\text{base/large}$~\citep{He2023DeBERTaV3}, and XLNet$_\text{base/large}$~\citep{Yang2019XLNet}, covering the encoder families previously shown effective for app-review NLP~\citep{Motger2024TFREX,Motger2025TFREXEMSE,Hou2024}.

\paragraph{Training setup.}
We use AdamW \citep{Loshchilov2019}, 3 epochs, batch size 8 and maximum sequence length 512. For the multi-label head, the epoch checkpoint with the highest macro-F1 on the evaluation fold is retained; binary classifiers use the final epoch.
The learning rate is tuned per backbone 
on the fixed tuning holdout of 152 sentences and shared across imbalance configurations for fair comparison (Table~\ref{tab:factors}).

\subsection{Decoder-only Classification}\label{sec:decoder-class}

\paragraph{Task formulation.}
Decoder-only models are evaluated as \emph{classifiers} (RQ2): they receive a review sentence, with its full review as context only, and return a multi-label emotion set.
The proprietary models are also used as \emph{generators} for synthetic augmentation (RQ3; Section~\ref{sec:data-imbalance}).

Classification prompts are built from the official annotation guidelines released with the ground truth~\citep{Motger2025Emotions}. These include taxonomy definitions, domain-specific emotion sections, disambiguation rules and annotation instructions. 
We evaluate:
\begin{enumerate}
\item \textbf{Zero-shot:} guideline definitions and annotation instructions, excluding the guideline's illustrative \texttt{[ExN]} examples.
\item \textbf{Few-shot (guidelines):} full guideline emotion sections, including those illustrative examples.
\item \textbf{Few-shot (guidelines + dataset):} the above plus up to five ground-truth sentences per emotion drawn from the training fold only.
\end{enumerate}

\paragraph{Model catalogue.}

We selected ten decoder-only LLMs as classifiers, organised in two cohorts: proprietary models accessed via API, and open-source models deployed locally (Table~\ref{tab:models}, decoder block).

For the proprietary cohort, we use popularity, accessibility for research, and fitness for large-scale multi-label emotion labelling as selection criteria.
With respect to popularity, we consider up-to-date statistical reports on AI chatbot market shares worldwide~\citep{statcounter2025chatbots}.
Concerning accessibility and experimental fitness, we require: (i)~API access suitable for batch classification and generation over the full labelled corpus and cross-validation folds; and (ii)~reliable support for structured (JSON) outputs.
For each selected provider, we selected a current conversational model available at the time of the experiments and suited to high-volume API use.
For OpenAI we use \texttt{gpt-5.2}, since its chat variants fix the decoding temperature.

For the open-source cohort, the selection followed three criteria: (1)~one instruction-tuned chat model per major open-source family; (2)~local deployability under our hardware, consisting of an NVIDIA GeForce RTX~4090 (24~GB VRAM), AMD Ryzen~9 7950X (16 cores), and 64~GB RAM on Ubuntu 24.04~LTS; and (3)~competitive standing among size-feasible variants on contemporary open LLM rankings such as the Artificial Analysis Intelligence Index as of July~2026~\citep{artificialanalysis2026}.
This resulted in six models from the Llama, Gemma, Qwen, OpenAI, Mistral, and DeepSeek families.
All open-source models were served locally through Ollama\footnote{\url{https://ollama.com/}} with the default quantisation of each tag.
Exact checkpoints, API identifiers and decoding parameters are recorded in the replication package.

\begin{table*}[t]
\centering
\caption{Large language models used in this study.
Encoder-only models are fine-tuned for classification (RQ1); the best backbone per formulation is also used in RQ4.
Decoder-only open-source models are used for zero-/few-shot classification (RQ2).
Proprietary decoder-only models are used for classification (RQ2) and as teachers for synthetic augmentation (RQ3). \texttt{mistral-large-2512} is released with open weights but is accessed through the provider's API and grouped with the proprietary cohort.}
\label{tab:models}
\footnotesize
\begin{tabular}{@{}llllll@{}}
\toprule
\textbf{Architecture} & \textbf{Access} & \textbf{Family} & \textbf{Variant} & \textbf{Identifier} & \textbf{Role} \\
\midrule
\multirow{9}{*}{Encoder-only}
 & Open-source & BERT & base & \texttt{bert-base-cased} & Fine-tune (RQ1, RQ4) \\
 & Open-source & BERT & large & \texttt{bert-large-cased} & Fine-tune (RQ1) \\
 & Open-source & RoBERTa & base & \texttt{roberta-base} & Fine-tune (RQ1) \\
 & Open-source & RoBERTa & large & \texttt{roberta-large} & Fine-tune (RQ1, RQ4) \\
 & Open-source & DistilBERT & base & \texttt{distilbert-base-cased} & Fine-tune (RQ1) \\
 & Open-source & DeBERTa-v3 & base & \texttt{microsoft/deberta-v3-base} & Fine-tune (RQ1) \\
 & Open-source & DeBERTa-v3 & large & \texttt{microsoft/deberta-v3-large} & Fine-tune (RQ1) \\
 & Open-source & XLNet & base & \texttt{xlnet-base-cased} & Fine-tune (RQ1) \\
 & Open-source & XLNet & large & \texttt{xlnet-large-cased} & Fine-tune (RQ1) \\
\midrule
\multirow{10}{*}{Decoder-only}
 & Open-source & DeepSeek & 8B & \texttt{deepseek-r1:8b} & Classify (RQ2) \\
 & Open-source & Mistral & 7B & \texttt{mistral:latest} & Classify (RQ2) \\
 & Open-source & OpenAI & 20B & \texttt{gpt-oss:20b} & Classify (RQ2) \\
 & Open-source & Gemma & 4B & \texttt{gemma3:4b} & Classify (RQ2) \\
 & Open-source & Llama & 8B & \texttt{llama3.1:8b} & Classify (RQ2) \\
 & Open-source & Qwen & 8B & \texttt{qwen3:8b} & Classify (RQ2) \\
 & Proprietary & OpenAI & 5.2 & \texttt{gpt-5.2} & Classify (RQ2) / Generate (RQ3) \\
 & Proprietary & Google & flash & \texttt{gemini-3.5-flash} & Classify (RQ2) / Generate (RQ3) \\
 & Proprietary & Anthropic & opus & \texttt{claude-opus-4-6} & Classify (RQ2) / Generate (RQ3) \\
 & Proprietary & Mistral & large & \texttt{mistral-large-2512} & Classify (RQ2) / Generate (RQ3) \\
\bottomrule
\end{tabular}
\end{table*}

\paragraph{Decoding setup.}
We use decoding temperature $T{=}0.0$ for primary comparisons and additionally evaluate the grid $T\in\{0.0, 0.3, 0.7\}$, shared across models and reported at each model's best prompting regime (Table~\ref{tab:factors}).

\paragraph{Output normalisation.}
Sentences are classified in batches of ten per request, and each response must return one labelled item per input identifier.
Model outputs are parsed into a subset of $E$. 
At most the first three valid labels are kept, and an empty set is scored as no prediction.

\subsection{Data Imbalance}\label{sec:data-imbalance}

\subsubsection{Loss-based methods}
Given our multi-label setting, we use binary cross-entropy on sigmoid logits~\citep{Tsoumakas2007,Cui2025Weighted}. Four loss-based methods are considered. For the binary ensemble, the same formulas are computed per binary classifier (i.e., with $|E|{=}1$).

\paragraph{(L1) BCEWithLogits + \texttt{pos\_weight}.}
For each emotion $c$, let $\mathrm{pos}_c$ and $\mathrm{neg}_c$ be the counts of positive and negative training instances in the current fold.
We set
\begin{equation}
\texttt{pos\_weight}[c]=\frac{\mathrm{neg}_c}{\mathrm{pos}_c},
\end{equation}
which upweights positive errors for rare emotions~\citep{Cui2025Weighted}.
This is our primary cost-sensitive baseline for multi-label training.

\paragraph{(L2) BCEWithLogits + class \texttt{weight}.}
As an alternative that scales both positive and negative contributions per class, we compute
\begin{equation}
w[c]=\frac{N}{|E|\cdot\mathrm{pos}_c},
\end{equation}
where $N$ is the number of training sentences in the fold~\citep{Cui2025Weighted}.
Implemented as a per-class multiplier on the BCE terms, this configuration is compared against (L1) to isolate the effect of positive-only versus class-level weighting.

\paragraph{(L3) Focal Loss + $\alpha$.}
Focal loss down-weights easy examples through the focusing parameter $\gamma$~\citep{Lin2017Focal} and has been applied to multi-label emotion classification~\citep{SemEval2025Focal}.
For multi-label data we use a sigmoid focal formulation with
\begin{equation}
\alpha[c]=\frac{\mathrm{neg}_c}{\mathrm{pos}_c+\mathrm{neg}_c}.
\end{equation}
We use the literature default $\gamma{=}2.0$~\citep{Lin2017Focal,SemEval2025Focal} with mean reduction (Table~\ref{tab:factors}).

\paragraph{(L4) Adaptive Focal Loss.}
Following~\citet{Labd2025AFL}, we evaluate Adaptive Focal Loss, which adjusts the focusing behaviour from class frequencies to reduce manual $\gamma$ tuning.

\subsubsection{Undersampling}
Within each training fold, sentences are selected iteratively, from the rarest to the most frequent emotion and in random order within each emotion, so that the positive count of each emotion does not exceed a cut-off $\tau$ while preserving minority positives.
We explore $\tau\in\{50,100,150,\ldots\}$ up to the majority-class size (i.e., \emph{Joy}, 371).

\subsubsection{Oversampling}
Let $M=\max_{c\in E}\mathrm{pos}_c$ be the majority label count in the human training data of a fold (or in the full training pool used for generation offline), where $E$ is the full nine-label set (Plutchik emotions plus \emph{Neutral}).
For each under-represented label $c\in E$, we generate synthetic reviews until $\mathrm{pos}_c$ reaches $M$.
Synthetic rows are stored separately from the ground truth and flagged with a \texttt{source=synthetic} attribute.
For fine-tuning, we inject up to $n$ synthetic positives per label in $E$, in bins of 50, to estimate the minimum augmentation needed for peak minority-class performance.

Synthetic generation is restricted to the proprietary models already selected for RQ2 classification (see Table~\ref{tab:models}).
This design follows recent work in which large language models generate labelled training data for compact task-specific models with orders of magnitude fewer parameters~\citep{Ye2022ZeroGen}; the diversity of the generated data is a key factor in downstream performance~\citep{Yu2023AttrPrompt}, and related distillation work shows that compact students can outperform a few-shot prompted large teacher~\citep{Hsieh2023Distilling}. Accordingly, these API models act as high-capacity teachers whose synthetic minority-emotion reviews transfer affective knowledge into compact encoder-only students for deployment.

Before large-scale generation, we run an augmentation-utility analysis on a balanced sample of synthetic reviews (20 per Plutchik emotion $\times$ strategy $\times$ generator) 
under three prompting strategies: zero-shot (definition only), few-shot with guideline examples, and few-shot with guideline examples plus five ground-truth reviews per emotion.
Reviews are embedded with \texttt{sentence-transformers/\allowbreak all-MiniLM-L6-v2}~\citep{Reimers2019,Wang2020MiniLM}.
For each label and configuration we compute:
\begin{itemize}
\item \texttt{diversity\_intra}: mean pairwise cosine distance within the synthetic set (higher $=$ broader coverage of the emotion spectrum);
\item \texttt{novelty\_vs\_synth}: mean nearest-neighbour distance within the synthetic set (higher $=$ fewer near-duplicates);
\item \texttt{novelty\_vs\_gt}: mean distance from each synthetic sample to its nearest human ground-truth neighbour of the same emotion (higher $=$ less paraphrasing of existing data);
\item \texttt{on\_emotion}: mean cosine similarity of synthetic samples to the human ground-truth centroid of the emotion (higher $=$ better label fidelity).
\end{itemize}
Configurations are ranked per label with Borda aggregation over the four metrics, and a Borda ranking over the label-averaged metrics yields a single recommended pipeline for generation across all labels in $E$.

The generation component exposes a target count of total positives per label (by default, majority-class parity) and stops when the quota is met.
Generated sentences are validated for non-emptiness and labelled with the requested target emotion.
When generative augmentation is active, each training fold contains the human fold subset plus the selected synthetic slice for that configuration (the first $n$ synthetic positives per label). 
For the binary ensemble, data-level methods are applied to the shared training pool of each fold.
Evaluation folds remain human-only.

To validate the selected generator extrinsically, three annotators label a stratified random sample of 100 synthetic reviews 
following the ground-truth annotation guidelines, blind to the target emotion. We report Fleiss' $\kappa$ and the precision, recall and F1 of the generator's target labels against the 2-of-3 majority vote.

\section{Results}\label{sec:results}

\subsection{Encoder-only Classification}\label{sec:encoder-class-res}

Table~\ref{tab:rq1-agg} reports aggregate performance for all nine encoder backbones under both formulations.
RoBERTa$_\text{large}$ is the best multi-label classifier (macro-F1 $0.387\pm0.032$), tied at three decimals with BERT$_\text{large}$ ($0.387\pm0.047$) but higher before rounding and with lower fold-to-fold variance, and ahead of RoBERTa$_\text{base}$ ($0.380\pm0.041$).
BERT$_\text{base}$ is best under the binary-ensemble formulation (macro-F1 $0.450\pm0.077$), ahead of BERT$_\text{large}$ ($0.440\pm0.063$) and RoBERTa$_\text{large}$ ($0.404\pm0.054$).
No backbone is simultaneously best under both formulations, although most backbones perform better under the binary-ensemble setting, with the exception of RoBERTa$_\text{base}$.
In fact, several models regress under multi-label, some sharply (e.g., down to macro-F1 $0.130$ for DeBERTa-v3$_\text{large}$, a $-0.25$ gap to its own binary-ensemble score).
Fold-to-fold variance (macro-F1 std) is larger for the binary ensemble on five of nine backbones.

\begin{table*}[t]
\centering
\caption{RQ1 aggregate performance across encoder backbones (10-fold CV, mean$\pm$std for macro-F1), listed as in Table~\ref{tab:models}. P/R: macro-averaged precision/recall. Bold marks the best backbone per formulation (by macro-F1).}
\label{tab:rq1-agg}
\small
\begin{tabular}{@{}lcccc cccc@{}}
\toprule
& \multicolumn{4}{c}{\textbf{Multi-label}} & \multicolumn{4}{c}{\textbf{Binary ensemble}} \\
\cmidrule(lr){2-5}\cmidrule(lr){6-9}
\textbf{Encoder} & \textbf{P} & \textbf{R} & \textbf{Macro-F1} & \textbf{Micro-F1} & \textbf{P} & \textbf{R} & \textbf{Macro-F1} & \textbf{Micro-F1} \\
\midrule
BERT$_\text{base}$ & 0.332 & 0.337 & 0.329$\pm$0.039 & 0.644 & 0.484 & 0.450 & \textbf{0.450$\pm$0.077} & 0.670 \\
BERT$_\text{large}$ & 0.417 & 0.383 & 0.387$\pm$0.047 & 0.671 & 0.463 & 0.448 & 0.440$\pm$0.063 & 0.673 \\
RoBERTa$_\text{base}$ & 0.405 & 0.381 & 0.380$\pm$0.041 & 0.683 & 0.326 & 0.348 & 0.327$\pm$0.052 & 0.641 \\
RoBERTa$_\text{large}$ & 0.410 & 0.390 & \textbf{0.387$\pm$0.032} & 0.697 & 0.413 & 0.415 & 0.404$\pm$0.054 & 0.696 \\
DistilBERT & 0.380 & 0.351 & 0.353$\pm$0.050 & 0.633 & 0.406 & 0.401 & 0.395$\pm$0.046 & 0.638 \\
DeBERTa-v3$_\text{base}$ & 0.201 & 0.250 & 0.218$\pm$0.044 & 0.572 & 0.286 & 0.310 & 0.286$\pm$0.063 & 0.619 \\
DeBERTa-v3$_\text{large}$ & 0.109 & 0.176 & 0.130$\pm$0.036 & 0.457 & 0.374 & 0.395 & 0.376$\pm$0.023 & 0.698 \\
XLNet$_\text{base}$ & 0.307 & 0.314 & 0.303$\pm$0.021 & 0.632 & 0.334 & 0.371 & 0.344$\pm$0.021 & 0.672 \\
XLNet$_\text{large}$ & 0.198 & 0.219 & 0.193$\pm$0.090 & 0.503 & 0.248 & 0.263 & 0.243$\pm$0.052 & 0.568 \\
\bottomrule
\end{tabular}
\end{table*}


Table~\ref{tab:rq1-per-emotion} breaks down per-emotion results for the best backbone under each formulation. 
Under the multi-label formulation, \emph{Joy}, \emph{Sadness}, \emph{Anticipation} and \emph{Trust} (support $\geq 152$) are classified strongly (F1 $0.69$--$0.80$), with poor performance on \emph{Neutral} and \emph{Disgust} (F1 $0.12$--$0.32$), while \emph{Fear}, \emph{Surprise} and \emph{Anger} are never predicted, with precision and recall both exactly zero.
Under the binary ensemble, no emotion is ever completely undetected: even its weakest (\emph{Surprise}, F1 $0.104$) has non-zero recall (discussed in Section~\ref{sec:findings}). 

\begin{table*}[t]
\centering
\caption{RQ1 per-emotion precision (P), recall (R) and F1 for the best backbone per formulation (RoBERTa$_\text{large}$, multi-label; BERT$_\text{base}$, binary ensemble). Support: positive sentences in the full 1{,}090-sentence corpus (encoder cross-validation uses the 938-sentence pool); sentences may carry multiple emotions. Per-emotion values are means over folds, so F1 need not equal the harmonic mean of the reported P and R. Sorted by support, descending.}
\label{tab:rq1-per-emotion}
\small
\begin{tabular}{@{}lrcccccc@{}}
\toprule
& & \multicolumn{3}{c}{\textbf{Multi-label}} & \multicolumn{3}{c}{\textbf{Binary ensemble}} \\
\cmidrule(lr){3-5}\cmidrule(lr){6-8}
\textbf{Emotion} & \textbf{Support} & \textbf{P} & \textbf{R} & \textbf{F1} & \textbf{P} & \textbf{R} & \textbf{F1} \\
\midrule
\emph{Joy} & 371 & 0.826 & 0.786 & 0.804 & 0.763 & 0.808 & 0.784 \\
\emph{Sadness} & 303 & 0.673 & 0.861 & 0.755 & 0.702 & 0.784 & 0.738 \\
\emph{Anticipation} & 181 & 0.798 & 0.794 & 0.791 & 0.728 & 0.820 & 0.765 \\
\emph{Trust} & 152 & 0.669 & 0.731 & 0.694 & 0.685 & 0.617 & 0.644 \\
\emph{Neutral} & 88 & 0.528 & 0.255 & 0.322 & 0.520 & 0.341 & 0.384 \\
\emph{Surprise} & 54 & 0.000 & 0.000 & 0.000 & 0.133 & 0.085 & 0.104 \\
\emph{Disgust} & 50 & 0.200 & 0.085 & 0.119 & 0.303 & 0.245 & 0.261 \\
\emph{Anger} & 27 & 0.000 & 0.000 & 0.000 & 0.350 & 0.150 & 0.207 \\
\emph{Fear} & 24 & 0.000 & 0.000 & 0.000 & 0.173 & 0.200 & 0.162 \\
\bottomrule
\end{tabular}
\end{table*}

Manual inspection of the multi-label classifier's false negatives on \emph{Anger}, \emph{Fear} and \emph{Surprise} shows a consistent pattern: in most instances, the classifier labels these reviews as \emph{Sadness} alone: 
e.g., \textit{``Very annoying encryption information takes up a lot of screen.''} (true: \emph{Anger}; predicted: \emph{Sadness}), \textit{``I mean, every aspect of your personal interactions become part of a terrifying data set that includes horrifying intrusions of privacy and information safety.''} (true: \emph{Fear}; predicted: \emph{Sadness}), and \textit{``I hate telegram''} (true: \emph{Anger}; predicted: \emph{Sadness}).
\emph{Sadness} is also the default on borderline \emph{Surprise} cases, e.g., \textit{``In Truecaller there is no video calling option...''} (true: \emph{Surprise}; predicted: \emph{Sadness}).


Table~\ref{tab:rq1-cost} reports inference latency per sentence for every backbone.

\begin{table}[htbp]
\centering
\caption{RQ1 inference latency per sentence (ms), over the 1{,}090 sentences with batch size 16 on the RTX~4090. Binary ensemble: nine times the latency of a single binary classifier.}
\label{tab:rq1-cost}
\small
\begin{tabular}{@{}lcc@{}}
\toprule
\textbf{Encoder} & \textbf{Multi-label (ms)} & \textbf{Binary ensemble (ms)} \\
\midrule
BERT$_\text{base}$ & 0.38 & 3.45 \\
BERT$_\text{large}$ & 1.10 & 9.90 \\
RoBERTa$_\text{base}$ & 0.38 & 3.43 \\
RoBERTa$_\text{large}$ & 1.08 & 9.74 \\
DistilBERT & 0.21 & 1.88 \\
DeBERTa-v3$_\text{base}$ & 0.72 & 4.73 \\
DeBERTa-v3$_\text{large}$ & 1.01 & 8.60 \\
XLNet$_\text{base}$ & 0.48 & 4.21 \\
XLNet$_\text{large}$ & 1.28 & 11.48 \\
\bottomrule
\end{tabular}
\end{table}

\subsection{Decoder-only Classification}\label{sec:decoder-class-res}

Table~\ref{tab:rq2-strategy} reports precision, recall, macro-F1 and micro-F1 for all ten decoders under each prompting strategy at $T{=}0$. 
Every decoder's best strategy is a few-shot variant; four (Qwen3-8B, DeepSeek-R1-8B, Mistral-7B, Gemma3-4B) peak at Few-shot (guidelines). 
Proprietary models dominate the top of the ranking: Gemini-3.5-Flash is best overall (macro-F1 $0.642$), ahead of GPT-5.2 ($0.577$) and Claude-Opus-4.6 ($0.544$).
Among open-source models, GPT-OSS-20B is the strongest (macro-F1 $0.526$), ahead of proprietary Mistral-Large ($0.504$) and clearly ahead of the remaining, smaller open-source models (macro-F1 $\leq 0.461$).
Only four decoders (DeepSeek-R1-8B, Llama3.1-8B, Mistral-7B, Gemma3-4B) trail the best encoder-only binary ensemble (BERT$_\text{base}$, macro-F1 $0.450$, Table~\ref{tab:rq1-agg}).
The best decoders (Gemini-3.5-Flash, GPT-5.2) clearly outperform every fine-tuned encoder on macro-F1, at far higher per-sentence latency: even the fastest decoder tested (Llama3.1-8B, $0.16$~s/sentence, Table~\ref{tab:rq2-cost}) is roughly $46$--$148\times$ slower than the RQ1 encoder configurations ($1.08$--$3.45$~ms/sentence, Table~\ref{tab:rq1-cost}), and Gemini-3.5-Flash's $1.21$~s/sentence is $351$--$1{,}120\times$ slower.
Proprietary decoders add per-request monetary cost on top, not included in our analysis.
The best-strategy precision/recall values show decoders are about as balanced as either RQ1 encoder formulation (Gemini-3.5-Flash: $0.637$/$0.685$, a $0.048$ gap, vs.\ RoBERTa$_\text{large}$'s $0.410$/$0.390$ and BERT$_\text{base}$ binary ensemble's $0.484$/$0.450$, Table~\ref{tab:rq1-agg}). However, Mistral-Large ($0.222$) and Llama3.1-8B ($0.141$) are the exceptions, with large precision--recall gaps.

\begin{table*}[t]
\centering
\caption{RQ2 precision (P), recall (R), macro-F1 and micro-F1 by decoder and prompting strategy ($T{=}0$). P/R: macro-averaged. Few-shot (guid.): guidelines; Few-shot (dataset): guidelines + dataset exemplars. Decoders are sorted by best macro-F1 within each cohort. Bold marks the best strategy (by macro-F1) per decoder.}
\label{tab:rq2-strategy}
\small
\begin{tabular}{@{}llcccc@{}}
\toprule
\textbf{Decoder} & \textbf{Strategy} & \textbf{P} & \textbf{R} & \textbf{Macro-F1} & \textbf{Micro-F1} \\
\midrule
\multicolumn{6}{@{}l}{\textit{Proprietary}} \\
\multirow{3}{*}{Gemini-3.5-Flash}
 & Zero-shot & 0.603 & 0.647 & 0.605 & 0.744 \\
 & Few-shot (guid.) & 0.595 & 0.638 & 0.594 & 0.753 \\
 & Few-shot (dataset) & 0.637 & 0.685 & \textbf{0.642} & 0.767 \\
\multirow{3}{*}{GPT-5.2}
 & Zero-shot & 0.538 & 0.575 & 0.528 & 0.669 \\
 & Few-shot (guid.) & 0.559 & 0.582 & 0.542 & 0.681 \\
 & Few-shot (dataset) & 0.587 & 0.623 & \textbf{0.577} & 0.699 \\
\multirow{3}{*}{Claude-Opus-4.6}
 & Zero-shot & 0.498 & 0.599 & 0.502 & 0.663 \\
 & Few-shot (guid.) & 0.522 & 0.595 & 0.525 & 0.682 \\
 & Few-shot (dataset) & 0.550 & 0.600 & \textbf{0.544} & 0.696 \\
\multirow{3}{*}{Mistral-Large}
 & Zero-shot & 0.456 & 0.686 & 0.497 & 0.609 \\
 & Few-shot (guid.) & 0.462 & 0.668 & 0.499 & 0.622 \\
 & Few-shot (dataset) & 0.464 & 0.686 & \textbf{0.504} & 0.621 \\
\multicolumn{6}{@{}l}{\textit{Open-source}} \\
\multirow{3}{*}{GPT-OSS-20B}
 & Zero-shot & 0.549 & 0.465 & 0.471 & 0.666 \\
 & Few-shot (guid.) & 0.575 & 0.489 & 0.500 & 0.702 \\
 & Few-shot (dataset) & 0.599 & 0.511 & \textbf{0.526} & 0.701 \\
\multirow{3}{*}{Qwen3-8B}
 & Zero-shot & 0.509 & 0.445 & 0.437 & 0.588 \\
 & Few-shot (guid.) & 0.500 & 0.459 & \textbf{0.461} & 0.639 \\
 & Few-shot (dataset) & 0.493 & 0.456 & 0.459 & 0.636 \\
\multirow{3}{*}{DeepSeek-R1-8B}
 & Zero-shot & 0.424 & 0.404 & 0.391 & 0.545 \\
 & Few-shot (guid.) & 0.463 & 0.452 & \textbf{0.436} & 0.602 \\
 & Few-shot (dataset) & 0.468 & 0.439 & 0.434 & 0.612 \\
\multirow{3}{*}{Llama3.1-8B}
 & Zero-shot & 0.315 & 0.447 & 0.321 & 0.417 \\
 & Few-shot (guid.) & 0.341 & 0.476 & 0.351 & 0.448 \\
 & Few-shot (dataset) & 0.381 & 0.522 & \textbf{0.389} & 0.485 \\
\multirow{3}{*}{Mistral-7B}
 & Zero-shot & 0.340 & 0.442 & 0.343 & 0.476 \\
 & Few-shot (guid.) & 0.368 & 0.463 & \textbf{0.368} & 0.504 \\
 & Few-shot (dataset) & 0.362 & 0.455 & 0.363 & 0.503 \\
\multirow{3}{*}{Gemma3-4B}
 & Zero-shot & 0.366 & 0.411 & 0.320 & 0.385 \\
 & Few-shot (guid.) & 0.398 & 0.428 & \textbf{0.361} & 0.434 \\
 & Few-shot (dataset) & 0.401 & 0.406 & 0.359 & 0.448 \\
\bottomrule
\end{tabular}
\end{table*}


Table~\ref{tab:rq2-temperature} reports temperature sensitivity for each decoder's best strategy.
All decoders are stable across the grid, from Claude-Opus-4.6 and Gemma3-4B (range $\leq 0.002$) to Llama3.1-8B (range $0.040$).
Temperature is therefore not a first-order factor for this task within the tested range, and $T{=}0$ is a reasonable default for reproducibility (the largest gain at $T{>}0$ is $+0.007$, Gemini-3.5-Flash at $T{=}0.3$).

\begin{table}[t]
\centering
\caption{RQ2 temperature sensitivity: macro-F1 at each decoder's best strategy. Decoders listed as in Table~\ref{tab:rq2-strategy}.}
\label{tab:rq2-temperature}
\small
\begin{tabular}{@{}lcccc@{}}
\toprule
\textbf{Decoder} & \textbf{$T{=}0$} & \textbf{$T{=}0.3$} & \textbf{$T{=}0.7$} & \textbf{Range} \\
\midrule
Gemini-3.5-Flash & 0.642 & 0.649 & 0.628 & 0.021 \\
GPT-5.2 & 0.577 & 0.573 & 0.570 & 0.007 \\
Claude-Opus-4.6 & 0.544 & 0.545 & 0.545 & 0.001 \\
Mistral-Large & 0.504 & 0.505 & 0.509 & 0.005 \\
GPT-OSS-20B & 0.526 & 0.508 & 0.532 & 0.024 \\
Qwen3-8B & 0.461 & 0.467 & 0.439 & 0.028 \\
DeepSeek-R1-8B & 0.436 & 0.421 & 0.427 & 0.015 \\
Llama3.1-8B & 0.389 & 0.378 & 0.349 & 0.040 \\
Mistral-7B & 0.368 & 0.366 & 0.363 & 0.005 \\
Gemma3-4B & 0.361 & 0.362 & 0.360 & 0.002 \\
\bottomrule
\end{tabular}
\end{table}

Table~\ref{tab:rq2-cost} reports the per-sentence cost profile of each decoder at its best strategy and $T{=}0$.
GPT-OSS-20B, Qwen3-8B and DeepSeek-R1-8B generate 102--145 completion tokens per sentence versus 19--31 for the proprietary models, and correspondingly show the highest latency alongside Gemini-3.5-Flash, whose latency likely reflects API response time (including network time) instead. 

\begin{table}[t]
\centering
\caption{RQ2 cost profile per sentence (best strategy, $T{=}0$, batched $b{=}10$). Decoders listed as in Table~\ref{tab:rq2-strategy}.}
\label{tab:rq2-cost}
\small
\begin{tabular}{@{}lccc@{}}
\toprule
\textbf{Decoder} & \textbf{Prompt tok.} & \textbf{Compl.\ tok.} & \textbf{Latency (s)} \\
\midrule
Gemini-3.5-Flash & 571 & 31 & 1.21 \\
GPT-5.2 & 540 & 19 & 0.23 \\
Claude-Opus-4.6 & 598 & 28 & 0.24 \\
Mistral-Large & 567 & 24 & 0.42 \\
GPT-OSS-20B & 542 & 130 & 0.80 \\
Qwen3-8B & 447 & 145 & 1.13 \\
DeepSeek-R1-8B & 450 & 102 & 0.80 \\
Llama3.1-8B & 546 & 20 & 0.16 \\
Mistral-7B & 508 & 28 & 0.19 \\
Gemma3-4B & 468 & 27 & 0.18 \\
\bottomrule
\end{tabular}
\end{table}

Table~\ref{tab:rq2-per-emotion} breaks down per-emotion results for the best proprietary and best open-source decoder. 
The same high-support/low-support pattern observed for encoders recurs here, but decoders recover \emph{Anger} far better than either encoder formulation (F1 $0.487$--$0.628$ vs.\ $0.000$--$0.207$).
\emph{Surprise} remains hard for both decoders (F1 $0.301$/$0.033$), echoing its weak performance under both encoder formulations, though neither decoder zeroes out the way the multi-label encoder does on \emph{Anger}, \emph{Fear} and \emph{Surprise} alike (Table~\ref{tab:rq1-per-emotion}).

\begin{table*}[t]
\centering
\caption{RQ2 per-emotion precision (P), recall (R) and F1 for the best proprietary (Gemini-3.5-Flash) and open-source (GPT-OSS-20B) decoder, each at its own best strategy, $T{=}0$. Support as in Table~\ref{tab:rq1-per-emotion}, sorted by support, descending.}
\label{tab:rq2-per-emotion}
\small
\begin{tabular}{@{}lrcccccc@{}}
\toprule
& & \multicolumn{3}{c}{\textbf{Gemini-3.5-Flash}} & \multicolumn{3}{c}{\textbf{GPT-OSS-20B}} \\
\cmidrule(lr){3-5}\cmidrule(lr){6-8}
\textbf{Emotion} & \textbf{Support} & \textbf{P} & \textbf{R} & \textbf{F1} & \textbf{P} & \textbf{R} & \textbf{F1} \\
\midrule
\emph{Joy} & 371 & 0.839 & 0.820 & 0.827 & 0.829 & 0.714 & 0.762 \\
\emph{Sadness} & 303 & 0.814 & 0.845 & 0.829 & 0.787 & 0.859 & 0.820 \\
\emph{Anticipation} & 181 & 0.803 & 0.895 & 0.844 & 0.788 & 0.773 & 0.771 \\
\emph{Trust} & 152 & 0.728 & 0.828 & 0.764 & 0.728 & 0.540 & 0.611 \\
\emph{Neutral} & 88 & 0.591 & 0.764 & 0.661 & 0.476 & 0.683 & 0.552 \\
\emph{Surprise} & 54 & 0.457 & 0.243 & 0.301 & 0.100 & 0.020 & 0.033 \\
\emph{Disgust} & 50 & 0.453 & 0.620 & 0.500 & 0.367 & 0.160 & 0.209 \\
\emph{Anger} & 27 & 0.575 & 0.750 & 0.628 & 0.700 & 0.400 & 0.487 \\
\emph{Fear} & 24 & 0.475 & 0.400 & 0.422 & 0.617 & 0.450 & 0.484 \\
\bottomrule
\end{tabular}
\end{table*}


\subsection{Data Imbalance}\label{sec:data-imbalance-res}

To select the generator and prompting strategy for synthetic review generation (RQ3), we ran an augmentation-utility analysis on a balanced sample of synthetic reviews per (strategy, generator) configuration, reported in full in Table~\ref{tab:rq3-ranking}.
Few-shot prompting (with or without dataset exemplars) outranks zero-shot within every generator, with \texttt{claude-opus-4-6} leading, \texttt{mistral-large-2512} second with guideline examples but tenth with dataset exemplars, and \texttt{gpt-5.2} trailing on all three of its strategies (ranks 8, 9 and 11).
The top configuration leads on \texttt{diversity\_intra}, \texttt{novelty\_vs\_synth} and \texttt{on\_emotion}, and separates most clearly on \texttt{novelty\_vs\_synth} ($0.539$ vs.\ $0.518$), while \texttt{diversity\_intra} varies least across the twelve configurations (range $0.086$ vs.\ $0.099$--$0.117$ for the other metrics). \texttt{novelty\_vs\_gt} runs against the other metrics, so the top configuration stays comparatively close to ground-truth phrasing.
We use the top-ranked configuration to build the synthetic pool for RQ4, which contains 2{,}218 reviews. 

\begin{table*}[t]
\centering
\caption{RQ3 augmentation-utility ranking across the four generators and three prompting strategies. \texttt{Div.}: \texttt{diversity\_intra}; \texttt{Nov.\ (synth)}: \texttt{novelty\_vs\_synth}; \texttt{Nov.\ (GT)}: \texttt{novelty\_vs\_gt}; \texttt{On-emo.}: \texttt{on\_emotion}; higher is better for all four metrics. Borda aggregates their per-metric ranks; overall rank sorts by Borda score, with the tie at ranks 11--12 broken by alphabetical order of the generator identifier. Bold marks the top-ranked configuration.}
\label{tab:rq3-ranking}
\small
\begin{tabular}{@{}clcccccc@{}}
\toprule
\textbf{Rank} & \textbf{Strategy} & \textbf{Generator} & \textbf{Div.} & \textbf{Nov.\ (synth)} & \textbf{Nov.\ (GT)} & \textbf{On-emo.} & \textbf{Borda} \\
\midrule
\textbf{1} & Few-shot (dataset) & \texttt{claude-opus-4-6} & 0.743 & 0.539 & 0.513 & 0.431 & \textbf{38} \\
2 & Few-shot (guidelines) & \texttt{mistral-large-2512} & 0.739 & 0.504 & 0.570 & 0.357 & 34 \\
3 & Few-shot (guidelines) & \texttt{claude-opus-4-6} & 0.727 & 0.518 & 0.511 & 0.429 & 33 \\
4 & Few-shot (dataset) & \texttt{gemini-3.5-flash} & 0.716 & 0.499 & 0.559 & 0.384 & 30 \\
5 & Few-shot (guidelines) & \texttt{gemini-3.5-flash} & 0.702 & 0.502 & 0.534 & 0.422 & 29 \\
6 & Zero-shot & \texttt{claude-opus-4-6} & 0.695 & 0.476 & 0.553 & 0.395 & 27 \\
7 & Zero-shot & \texttt{gemini-3.5-flash} & 0.689 & 0.470 & 0.589 & 0.349 & 24 \\
8 & Few-shot (guidelines) & \texttt{gpt-5.2} & 0.680 & 0.452 & 0.550 & 0.407 & 22 \\
9 & Few-shot (dataset) & \texttt{gpt-5.2} & 0.666 & 0.438 & 0.549 & 0.409 & 20 \\
10 & Few-shot (dataset) & \texttt{mistral-large-2512} & 0.706 & 0.422 & 0.544 & 0.392 & 19 \\
11 & Zero-shot & \texttt{gpt-5.2} & 0.657 & 0.436 & 0.610 & 0.330 & 18 \\
12 & Zero-shot & \texttt{mistral-large-2512} & 0.681 & 0.436 & 0.591 & 0.328 & 18 \\
\bottomrule
\end{tabular}
\end{table*}

In the human-as-judge validation (Section~\ref{sec:data-imbalance}), three annotators (authors of this study, one of whom had not taken part in the original ground-truth annotation) labelled 100 synthetic reviews.
Pooled across all nine emotions, agreement is substantial (Fleiss' $\kappa=0.749$). 
Against the human majority vote, the generator's assigned labels reach precision~$=0.950$, recall~$=0.941$, and F1~$=0.945$.
Five of the nine emotions (i.e., \emph{Fear}, \emph{Sadness}, \emph{Anger}, \emph{Anticipation} and \emph{Neutral}) reach perfect scores. The remaining four (\emph{Joy}, \emph{Trust}, \emph{Surprise}, \emph{Disgust}) account for all of the residual error, aligning with originally reported challenges in emotion annotation for those emotions~\citep{Motger2025Emotions}.
This high fidelity supports the RQ3 selection of this generator and strategy.
Full annotations and reported metrics are available in the replication package.

Imbalance mitigation (RQ4) was evaluated on the best RQ1 backbone per formulation (RoBERTa$_\text{large}$, multi-label; BERT$_\text{base}$, binary ensemble), sweeping loss-reweighting, undersampling and generative augmentation individually and in nine data-level $\times$ loss-reweighting (L1--L3) combinations (Table~\ref{tab:rq4-sweep}).

\begin{table*}[htbp]
\centering
\caption{RQ4 imbalance mitigation sweep on the best RQ1 backbone per formulation (RoBERTa$_\text{large}$, multi-label; BERT$_\text{base}$, binary ensemble; 10-fold CV, mean$\pm$std; P/R: macro-averaged). L1--L4: loss reweighting; $\tau$: undersampling cut-off; $n$: injected synthetic reviews per emotion. 
Bold marks the best configuration per formulation (by macro-F1).}
\label{tab:rq4-sweep}
\footnotesize
\begin{tabular}{@{}lcccc cccc@{}}
\toprule
& \multicolumn{4}{c}{\textbf{Multi-label}} & \multicolumn{4}{c}{\textbf{Binary ensemble}} \\
\cmidrule(lr){2-5}\cmidrule(lr){6-9}
\textbf{Configuration} & \textbf{P} & \textbf{R} & \textbf{Macro-F1} & \textbf{Micro-F1} & \textbf{P} & \textbf{R} & \textbf{Macro-F1} & \textbf{Micro-F1} \\
\midrule
Baseline (none) & 0.410 & 0.390 & 0.387$\pm$0.032 & 0.697 & 0.484 & 0.450 & 0.450$\pm$0.077 & 0.670 \\
\multicolumn{9}{@{}l}{\textit{Loss reweighting}} \\
BCE + pos.\ weight (L1) & 0.242 & 0.527 & 0.303$\pm$0.151 & 0.443 & 0.361 & 0.378 & 0.359$\pm$0.044 & 0.632 \\
BCE + class weight (L2) & 0.206 & 0.214 & 0.185$\pm$0.120 & 0.476 & 0.355 & 0.360 & 0.345$\pm$0.023 & 0.635 \\
Focal loss (L3) & 0.396 & 0.728 & 0.462$\pm$0.021 & 0.553 & 0.442 & 0.551 & 0.461$\pm$0.039 & 0.606 \\
Adaptive focal loss (L4) & 0.308 & 0.603 & 0.358$\pm$0.090 & 0.445 & 0.452 & 0.574 & 0.467$\pm$0.042 & 0.593 \\
\multicolumn{9}{@{}l}{\textit{Undersampling ($\tau$ positives/emotion)}} \\
$\tau{=}50$ & 0.268 & 0.273 & 0.246$\pm$0.092 & 0.438 & 0.412 & 0.394 & 0.373$\pm$0.046 & 0.513 \\
$\tau{=}100$ & 0.358 & 0.366 & 0.338$\pm$0.033 & 0.613 & 0.370 & 0.383 & 0.362$\pm$0.051 & 0.584 \\
$\tau{=}150$ & 0.360 & 0.377 & 0.360$\pm$0.038 & 0.663 & 0.415 & 0.413 & 0.395$\pm$0.048 & 0.645 \\
$\tau{=}200$ & 0.421 & 0.396 & 0.394$\pm$0.041 & 0.699 & 0.444 & 0.421 & 0.414$\pm$0.063 & 0.655 \\
$\tau{=}250$ & 0.387 & 0.383 & 0.376$\pm$0.029 & 0.690 & 0.415 & 0.411 & 0.402$\pm$0.048 & 0.655 \\
$\tau{=}300$ & 0.399 & 0.387 & 0.381$\pm$0.029 & 0.698 & 0.365 & 0.368 & 0.353$\pm$0.043 & 0.635 \\
$\tau{=}350$ & 0.408 & 0.389 & 0.386$\pm$0.032 & 0.696 & 0.366 & 0.369 & 0.354$\pm$0.042 & 0.636 \\
\multicolumn{9}{@{}l}{\textit{Generative augmentation ($n$ injected/emotion)}} \\
$n{=}50$ & 0.585 & 0.533 & 0.541$\pm$0.091 & 0.730 & 0.490 & 0.475 & 0.466$\pm$0.055 & 0.651 \\
$n{=}100$ & 0.614 & 0.551 & 0.560$\pm$0.049 & 0.726 & 0.518 & 0.482 & 0.483$\pm$0.079 & 0.652 \\
$n{=}150$ & 0.631 & 0.546 & 0.564$\pm$0.050 & 0.726 & 0.540 & 0.503 & 0.502$\pm$0.074 & 0.673 \\
$n{=}200$ & 0.631 & 0.571 & 0.580$\pm$0.087 & 0.740 & 0.522 & 0.516 & 0.498$\pm$0.084 & 0.659 \\
$n{=}250$ & 0.627 & 0.554 & 0.564$\pm$0.056 & 0.721 & 0.546 & 0.526 & 0.516$\pm$0.087 & 0.677 \\
$n{=}300$ & 0.626 & 0.556 & 0.570$\pm$0.058 & 0.727 & 0.538 & 0.515 & 0.510$\pm$0.073 & 0.664 \\
$n{=}350$ & 0.610 & 0.547 & 0.556$\pm$0.070 & 0.706 & 0.486 & 0.486 & 0.474$\pm$0.076 & 0.647 \\
\multicolumn{9}{@{}l}{\textit{Combined}} \\
Undersample ($\tau{=}150$) + L1 & 0.294 & 0.649 & 0.374$\pm$0.143 & 0.494 & 0.374 & 0.419 & 0.387$\pm$0.032 & 0.630 \\
Undersample ($\tau{=}150$) + L2 & 0.290 & 0.293 & 0.275$\pm$0.071 & 0.570 & 0.363 & 0.398 & 0.368$\pm$0.027 & 0.641 \\
Undersample ($\tau{=}150$) + L3 & 0.338 & 0.679 & 0.410$\pm$0.081 & 0.498 & 0.456 & 0.543 & 0.460$\pm$0.062 & 0.565 \\
GenAI aug ($n{=}100$) + L1 & 0.566 & 0.698 & \textbf{0.591$\pm$0.054} & 0.700 & 0.541 & 0.520 & 0.510$\pm$0.083 & 0.664 \\
GenAI aug ($n{=}100$) + L2 & 0.616 & 0.544 & 0.561$\pm$0.079 & 0.723 & 0.527 & 0.467 & 0.475$\pm$0.074 & 0.663 \\
GenAI aug ($n{=}100$) + L3 & 0.504 & 0.756 & 0.582$\pm$0.056 & 0.687 & 0.536 & 0.565 & \textbf{0.530$\pm$0.074} & 0.669 \\
Undersample + GenAI aug + L1 & 0.527 & 0.692 & 0.569$\pm$0.049 & 0.680 & 0.553 & 0.535 & 0.517$\pm$0.040 & 0.651 \\
Undersample + GenAI aug + L2 & 0.592 & 0.554 & 0.553$\pm$0.066 & 0.713 & 0.514 & 0.504 & 0.482$\pm$0.077 & 0.640 \\
Undersample + GenAI aug + L3 & 0.482 & 0.734 & 0.558$\pm$0.052 & 0.648 & 0.519 & 0.571 & 0.525$\pm$0.086 & 0.659 \\
\bottomrule
\end{tabular}
\end{table*}

Loss reweighting alone is unreliable: only focal loss (L3) improves over the multi-label baseline (macro-F1 $0.462$ vs.\ $0.387$); L1, L2 and L4 all make it worse.
The two BCE variants fail for opposite reasons, and both are consistent with the calibration mechanism hypothesised for RQ1 (Section~\ref{sec:findings}).
Positive weighting pushes predicted probabilities up so aggressively that most sentences hit the $k{=}3$ label cap regardless of evidence, collapsing precision ($0.242$).
Class weighting pushes them down instead, so most predictions fall back to a single guessed label, collapsing both recall ($0.214$) and precision ($0.206$).
Focal loss also raises predicted probabilities above baseline, but it may preserve which labels are genuinely more likely, so precision holds much better ($0.396$) even as recall improves, at a cost in micro-F1 ($0.553$ vs.\ $0.697$).
The binary ensemble is far less sensitive to this than the multi-label head under L2--L4 (under L1 both drop similarly: $-0.091$ vs.\ $-0.084$), consistent with our hypothesis (Section~\ref{sec:findings}) that independent per-emotion classifiers calibrate more reliably than one shared, jointly-trained head.

\begin{table*}[t]
\centering
\caption{RQ4 per-emotion precision (P), recall (R) and F1, baseline vs.\ best configuration (RoBERTa$_\text{large}$, multi-label; GenAI aug.\ $n{=}100$ + BCE positive weight). Support as in Table~\ref{tab:rq1-per-emotion}, sorted by support, descending.}
\label{tab:rq4-per-emotion}
\small
\begin{tabular}{@{}lrccccccc@{}}
\toprule
& & \multicolumn{3}{c}{\textbf{Baseline}} & \multicolumn{3}{c}{\textbf{Best config.}} & \\
\cmidrule(lr){3-5}\cmidrule(lr){6-8}
\textbf{Emotion} & \textbf{Support} & \textbf{P} & \textbf{R} & \textbf{F1} & \textbf{P} & \textbf{R} & \textbf{F1} & \textbf{$\Delta$F1} \\
\midrule
\emph{Joy} & 371 & 0.826 & 0.786 & 0.804 & 0.750 & 0.878 & 0.806 & +0.002 \\
\emph{Sadness} & 303 & 0.673 & 0.861 & 0.755 & 0.713 & 0.842 & 0.771 & +0.016 \\
\emph{Anticipation} & 181 & 0.798 & 0.794 & 0.791 & 0.764 & 0.885 & 0.815 & +0.024 \\
\emph{Trust} & 152 & 0.669 & 0.731 & 0.694 & 0.639 & 0.847 & 0.720 & +0.026 \\
\emph{Neutral} & 88 & 0.528 & 0.255 & 0.322 & 0.537 & 0.538 & 0.507 & +0.185 \\
\emph{Surprise} & 54 & 0.000 & 0.000 & 0.000 & 0.335 & 0.495 & 0.346 & +0.346 \\
\emph{Disgust} & 50 & 0.200 & 0.085 & 0.119 & 0.374 & 0.630 & 0.448 & +0.329 \\
\emph{Anger} & 27 & 0.000 & 0.000 & 0.000 & 0.545 & 0.667 & 0.501 & +0.501 \\
\emph{Fear} & 24 & 0.000 & 0.000 & 0.000 & 0.435 & 0.500 & 0.410 & +0.410 \\
\bottomrule
\end{tabular}
\end{table*}

Undersampling stays near or below baseline across every cut-off (only $\tau{=}200$ marginally exceeds it for multi-label: $0.394$ vs.\ $0.387$), while generative augmentation provides promising results.
For multi-label, every tested injection size beats the baseline by a wide margin ($n{=}50$: $0.541$; peaking at $n{=}200$: $0.580$), consistent with baseline's complete failure to detect \emph{Fear}, \emph{Surprise} and \emph{Anger} (Table~\ref{tab:rq1-per-emotion}).
Unlike the reweighting losses, augmentation keeps the model well-calibrated ($0.99$ threshold crossings per sentence on average, against a ground-truth mean of $1.15$ labels).

Pairing each data-level method with each of L1--L3 (at $\tau{=}150$ and $n{=}100$, taken as representative settings from the standalone sweeps) confirms that no single reweighting scheme is uniformly the best combination partner.
Undersampling still pairs best with focal loss (L3: $0.410$, vs.\ $0.374$ for L1 and $0.275$ for L2), but generative augmentation pairs best with positive weighting (L1: $0.591$), narrowly ahead of focal (L3: $0.582$, within one standard deviation) and clearly ahead of class weighting (L2: $0.561$).
Combining both data-level methods together pairs best with L1 too ($0.569$) yet still falls short of augmentation alone paired with L1, so stacking undersampling on top of augmentation does not help further.
The overall best multi-label configuration is generative augmentation ($n{=}100$) with BCE positive weighting (macro-F1 $0.591\pm0.054$, $+0.204$ over baseline).
L1's own standalone score ($0.303$) trails L3's ($0.462$) by a wide margin, so this combination winner is not predictable from standalone loss rankings alone.
Calibration is also consistent with L1 overtaking L3 only once augmentation is added: augmentation supplies enough genuine minority-class positives that positive weighting's over-triggering is largely corrected ($3.15\to1.49$ average threshold crossings once paired with $n{=}100$ synthetic reviews), whereas focal loss's own, milder over-triggering is corrected less completely under the same pairing ($2.68\to1.61$).

For the binary ensemble, the best configuration is generative augmentation paired with focal loss (L3: $0.530\pm0.074$, $+0.080$ over its own baseline).
For multi-label, the gap to the best decoder (Gemini-3.5-Flash, macro-F1 $0.642$, Table~\ref{tab:rq2-strategy}) shrinks from $0.255$ at the RQ1 baseline to just $0.051$ after mitigation.
Binary ensemble narrows its own gap by less: from $0.192$ at baseline to $0.112$ after mitigation.
Either way, the cost advantage from RQ1 is untouched, since mitigation changes only the training data and loss, not the architecture used at inference. 

Table~\ref{tab:rq4-per-emotion} contrasts per-emotion F1 between the baseline and the best configuration. 
This configuration rescues the three emotions the baseline never predicts: \emph{Anger} goes from F1 $0$ to $0.501$, \emph{Fear} from $0$ to $0.410$, and \emph{Surprise} from $0$ to $0.346$, with \emph{Disgust} close behind ($0.119\rightarrow 0.448$).
Furthermore, \emph{Joy}, \emph{Sadness}, \emph{Anticipation} and \emph{Trust} all improve slightly ($+0.002$ to $+0.026$).

\begin{figure*}[htbp]
\centering
\includegraphics[width=0.9\textwidth]{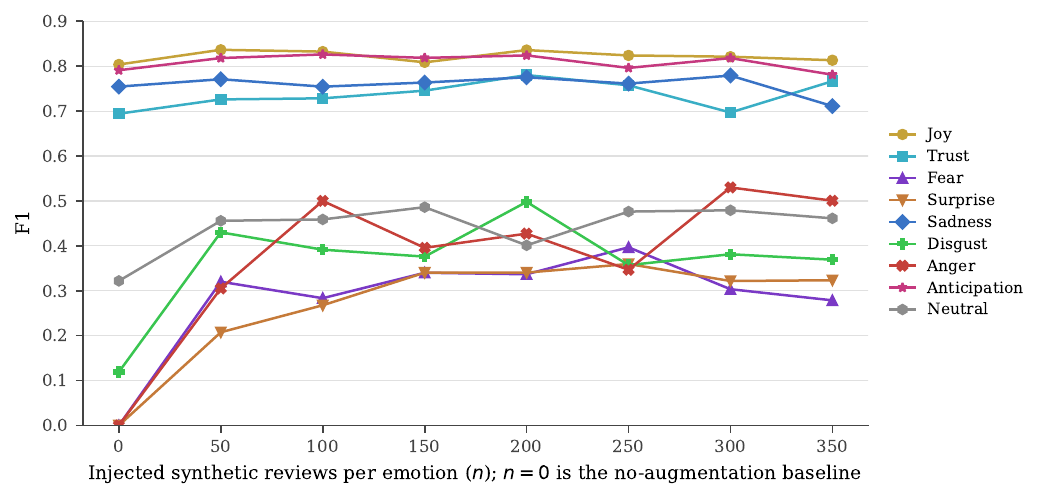}
\caption{RQ4: per-emotion F1 as a function of the generative-augmentation injection size $n$ (RoBERTa$_\text{large}$, multi-label; pure augmentation, no reweighting); $n{=}0$ is the no-mitigation baseline.}
\label{fig:rq4-per-emotion}
\end{figure*}

Standalone augmentation's aggregate macro-F1 levels off across $n{=}50$ to $n{=}350$ (Table~\ref{tab:rq4-sweep}); Figure~\ref{fig:rq4-per-emotion} disaggregates this trajectory by emotion.
\emph{Fear}, \emph{Surprise}, \emph{Anger} and \emph{Disgust} all make their largest single jump at the smallest tested injection (\emph{Fear}: $0\!\rightarrow\!0.320$; \emph{Surprise}: $0\!\rightarrow\!0.207$; \emph{Anger}: $0\!\rightarrow\!0.305$; \emph{Disgust}: $0.119\!\rightarrow\!0.430$, all at $n{=}50$), confirming that even modest augmentation is enough to break the zero-detection failure mode identified in Table~\ref{tab:rq1-per-emotion}.
Beyond $n{=}50$ these curves are not monotonic: for example, \emph{Anger}'s best point is $n{=}300$ ($0.530$), and \emph{Fear}'s is also $n{=}300$ ($0.397$) but it dips again by $n{=}350$.
This flat aggregate trend reflects that the first injection bin does most of the work.
\emph{Neutral} follows the same early-jump pattern ($0.322\!\rightarrow\!0.456$ at $n{=}50$) with one mid-sweep dip at $n{=}250$ ($0.401$).
\emph{Trust}, \emph{Sadness}, \emph{Joy} and \emph{Anticipation} stay within a narrow band throughout (at most $\pm0.05$), as expected for classes that were never data-starved.
Each point is a single 10-fold CV mean, so the injection size at which any one emotion peaks should be read qualitatively.


\section{Discussion}\label{sec:discussion}

\subsection{Findings}\label{sec:findings}

\begin{rqbox}{RQ1 --- Encoder-only classification}
Fine-tuned encoders reach moderate performance (best macro-F1 $0.387$ multi-label, RoBERTa$_\text{large}$; $0.450$ binary ensemble, BERT$_\text{base}$).
Binary ensemble is the better formulation for eight of nine backbones, and the multi-label head has a complete blind spot on \emph{Fear}, \emph{Surprise} and \emph{Anger} that binary ensemble does not share.
\end{rqbox}

Overall, macro-F1 remains modest across all nine backbones, which is consistent with the task (i.e., fine-grained, multi-label, and severely imbalanced emotion classification on short, informal app-review sentences).
The backbone dependence in Table~\ref{tab:rq1-agg} may reflect a calibration issue, as both formulations use the same threshold-based decision rule. Hence, a backbone that ranks the correct labels above the incorrect ones can still score badly if its predicted probabilities are not meaningful in absolute terms.
Binary ensemble's independently-trained classifiers appear more robust to this than the multi-label head's shared, jointly-trained representation. This is also the more consequential explanation for the multi-label head's complete blind spot on \emph{Fear}, \emph{Surprise} and \emph{Anger} (Table~\ref{tab:rq1-per-emotion}), given that a jointly-trained head may have less room to accommodate a badly-calibrated emotion than nine independently-trained classifiers.

The relabelling-to-\emph{Sadness} pattern behind that blind spot (Table~\ref{tab:rq1-per-emotion}) echoes the annotation difficulty already reported for this ground truth: semantically overlapping emotions with shared cues that trained annotators themselves disagreed on (Challenge~1), and is compounded here by sentence-level labelling without the fuller review context that could disambiguate them (Challenge~4)~\citep{Motger2025Emotions}.
Comparing predicted label-cardinality distributions against the ground truth's ($85.8\%$ one emotion, $13.8\%$ two, $0.5\%$ three) shows binary ensemble's threshold rule lands close to the true distribution regardless of configuration (e.g., $87\%$/$13\%$/$0\%$ at baseline), while the multi-label head swings from under-triggering at baseline ($95\%$/$5\%$/$0\%$) to over-triggering once augmented and positive-weighted ($58\%$/$36\%$/$6\%$, Section~\ref{sec:data-imbalance-res}).
Macro-F1 on rare emotions is therefore not purely a data-volume problem solvable by more positives alone. Part of the ceiling is the same semantic overlap that makes these labels hard for humans, which augmentation and reweighting rebalance partially (Section~\ref{sec:data-imbalance-res}), but do not remove entirely.

\begin{rqbox}{RQ2 --- Decoder-only classification}
Few-shot prompting yields the best strategy for every decoder. The best proprietary model (Gemini-3.5-Flash, few-shot with dataset exemplars, macro-F1 $0.642$) outperforms every fine-tuned encoder, and the best open-source local model (GPT-OSS-20B, macro-F1 $0.526$) is competitive with mid-tier proprietary models.
\end{rqbox}

The fact that guideline-only exemplars already suffice for four of the ten decoders (Table~\ref{tab:rq2-strategy}) suggests that, for these smaller models, additional dataset exemplars mainly add prompt length without measurable gain (differences $\leq 0.005$), whereas the other six decoders gain up to $+0.048$.
The cost asymmetry behind RQ2's headline result (Tables~\ref{tab:rq1-cost} and~\ref{tab:rq2-cost}) is a significant trade-off, revisited in RQ4: no fine-tuned encoder matches the best decoder on macro-F1, but every decoder is one to three orders of magnitude slower. 


\begin{rqbox}{RQ3 --- Selecting and configuring generative augmentation}
Few-shot prompting seeded with the annotation guidelines' own illustrative examples plus dataset exemplars, generated by \texttt{claude-opus-4-6}, produces the synthetic reviews with the best diversity/novelty/fidelity profile, and the selected configuration yields the downstream gains observed under RQ4.
\end{rqbox}

The augmentation-utility ranking is computed without training a classifier, on a small balanced sample (20 reviews per label, strategy and generator) with a single sentence-embedding model. It should therefore be read as a heuristic filter: it captures corpus-level properties of the synthetic pool but does not certify that any individual generated sentence carries the correct label. Human assessment indicates that the selected LLM configuration can be used for data augmentation, as long as proper context engineering practices (e.g., annotation guidelines, data samples) are adopted. Nevertheless, human oversight remains essential for ambiguous emotions that are difficult to annotate, in contrast with emotions showing perfect correctness in generation. Since only the top-ranked configuration was injected in RQ4, whether the ranking itself predicts downstream usefulness remains untested.

\begin{rqbox}{RQ4 --- Data imbalance mitigation}
Generative augmentation is the only strategy that improves on the baseline at every tested injection size and for both formulations. Pairing it with BCE positive-weighted loss yields the best multi-label configuration overall (macro-F1 $0.591\pm0.054$, $+0.204$ over baseline), recovering all three previously undetected emotions with no majority-class F1 regression, while loss reweighting and undersampling remain unreliable in isolation.
This narrows, but does not close, the effectiveness gap to RQ2's best decoder (macro-F1 $0.591$ vs.\ $0.642$, a $0.051$ deficit) while keeping the mitigated encoder's two-to-three-orders-of-magnitude inference-latency advantage intact.
\end{rqbox}

Only focal loss (L3) improves over the multi-label baseline, while both focal variants do so for binary ensemble (L3: $0.461$; L4: $0.467$).
Both BCE variants are consistent with the calibration mechanism hypothesised above for RQ1, from opposite directions (Section~\ref{sec:data-imbalance-res}).
Undersampling remains unreliable in isolation: every binary-ensemble cut-off remains below its baseline ($0.450$). 
Generative augmentation's benefit is asymmetric across formulations: averaged over injection sizes, it delivers about four times the multi-label gain it delivers for binary ensemble ($+0.175$ vs.\ $+0.043$, Table~\ref{tab:rq4-sweep}). Presumably, augmentation mitigates the blind spot problem in the multi-label classification head, while binary ensemble already predicted every emotion at least sometimes and had comparatively little headroom to gain.

Which reweighting loss pairs best with which data-level method is unpredictable from standalone performance (Section~\ref{sec:data-imbalance-res}).
Practitioners facing similar imbalance issues should not pick the best-looking loss in isolation and assume it carries over to a paired configuration: the design space has to be swept empirically, a modest one-off cost against the risk of deploying a badly-paired configuration.
Since mitigation narrows the gap to RQ2's best decoder much more for multi-label than for binary ensemble, while leaving inference cost unchanged, the preferred setting is an application-level throughput/cost-vs.-accuracy decision.

Mitigation strategies themselves are not equally cheap to obtain. Loss reweighting and undersampling change only a training-time hyperparameter---undersampling, if anything, trains on \emph{fewer} sentences---whereas the augmentation pipeline behind the largest gains required generating $2{,}218$ synthetic reviews across the nine emotions through proprietary LLM APIs (up to $n{=}350$ per emotion, Section~\ref{sec:data-imbalance}), each incurring per-request monetary and latency cost before a single classifier is trained.
The accuracy gains may justify this one-off generation cost, reused across every downstream configuration, but the cheaper options are not strictly dominated: undersampling at $\tau{=}200$ is nearly free and, for multi-label, marginally exceeds the baseline, which may suit resource-constrained deployments.

\subsection{Threats to Validity}\label{sec:ttv}

We organise threats to validity following~\citet{Wohlin2012}.

\paragraph{Conclusion validity.}
Encoder results (RQ1, RQ4) are averaged over 10-fold cross-validation with reported standard deviations, but decoder results (RQ2) and the RQ3 augmentation-utility ranking are each a single pass over the test set or sample at a fixed temperature; point differences below roughly $0.02$ macro-F1 for decoders, or between closely-ranked configurations in Table~\ref{tab:rq3-ranking}, should be read as indicative.
Rare-emotion F1 (support: \emph{Fear} 24; \emph{Anger} 27; \emph{Disgust} 50; \emph{Surprise} 54) is estimated from very few positive instances per fold, which is reflected in the large standard deviations for several RQ4 configurations (up to $\pm0.15$ macro-F1, Table~\ref{tab:rq4-sweep}). Claims about ``rescuing'' these emotions rest on a comparatively small evidence base.

\paragraph{Internal validity.}
The cap $k{=}3$ of the threshold-then-cap decision rule ($\theta{=}0.5$) was itself derived from the full ground truth's own label-cardinality distribution (Section~\ref{sec:dataset}); a different rule could change which formulation wins for a given backbone (Section~\ref{sec:findings}), so the multi-label-vs-binary-ensemble comparison is conditioned on this specific design choice. 
Encoder learning-rate tuning (Table~\ref{tab:factors}) was performed once per backbone and reused across the full RQ4 sweep, so RQ4 results may understate what a configuration-specific tuning pass could achieve; epochs and batch size are fixed for all backbones, which may under-train the large variants.
For the multi-label head, the retained epoch checkpoint is selected on the evaluation fold itself (with a top-3 assignment rule), which may bias multi-label results upwards; injection sizes, cut-offs and the reported best configurations are likewise selected on cross-validated results rather than on an independent split.
Decoders receive the full review as context, whereas encoders see only the sentence, so the RQ1--RQ2 comparison also reflects this input asymmetry.
The public release of the ground truth and guidelines before the experiments means that the proprietary decoders and generators may have seen them during pre-training.
The human-as-judge annotators are authors of this study, two of whom also produced the original ground truth.

\paragraph{Construct validity.}
Macro-F1 under a fixed $\theta{=}0.5$ threshold operationalises ``correct classification'' in one specific way. The RQ1 interpretation that some backbones may rank labels correctly but still score badly under this threshold (Section~\ref{sec:findings}) suggests that a differently-operationalised construct, e.g., a ranking metric or a per-emotion calibrated threshold, could shift some conclusions.
Plutchik's eight emotions plus \emph{Neutral}~\citep{Plutchik2001} is one validated but not exhaustive taxonomy, and annotating at sentence level without the surrounding review is itself a construct simplification, one the source ground truth's own Challenge~4 already documents~\citep{Motger2025Emotions}.
Furthermore, whether a generated review's assigned emotion is genuinely expressed is assessed by the human-as-judge validation (Section~\ref{sec:data-imbalance-res}) on a sample of 100 reviews only. The augmentation-utility metrics depend on a single general-purpose embedding model: \texttt{on\_emotion} captures topical similarity to same-emotion ground truth rather than emotion fidelity as such, and synthetic items are single-label reviews added to a sentence-level multi-label corpus.

\paragraph{External validity.}
All experiments use a single, English-language, Google-Play-only ground truth~\citep{Motger2025Emotions}; conclusions may not transfer to other app stores, languages, or annotation taxonomies without re-validation.
Decoder and generator comparisons reflect specific model snapshots at experiment time. Both capability and cost change quickly, so absolute numbers, some relative rankings, and the latency and pricing gap behind the cost-effectiveness argument may not hold for later model generations.

\section{Related Work}\label{sec:rel-work}

LLM-based emotion extraction has gained significant attention across heterogeneous domains, including social media~\citep{AMEER2023118534} and psychotherapy~\citep{Lalk2025}. Evidence shows that LLMs encode a low-dimensional emotional knowledge that generalises across emotions, datasets, languages and styles, supporting the view that emotion signals are (at least) partly transferable across domains~\citep{reichman2025emotions}. This motivates domain-specific emotion extraction built on pre-trained foundation models, the setting of this study.

At the same time, domain transfer remains fragile when training data and target emotions differ in style and semantic interpretation. For example, in psychotherapy transcripts, fine-tuned LLMs achieve only moderate macro-F1, attributed to short, informal, noisy training data and poor balance for rare emotions~\citep{Lalk2025}. We address the same weaknesses by grounding the task in app-review-specific labels, multi-label supervision, guidelines adoption, and domain-specific imbalance mitigation strategies, rather than assuming cross-domain robustness.

In software engineering communication, general-purpose emotion classifiers suffer substantial performance degradation when applied across channels (e.g., GitHub, Stack Overflow, Jira), which the authors attribute to domain-specific style and vocabulary mismatches rather than model weakness, motivating targeted data augmentation over reliance on off-the-shelf, cross-domain models~\citep{Imran2022}.
While several datasets have been used, only a few are publicly available (see Section~\ref{sec:emo-datasets-b}). Furthermore, automated extraction remains very limited within the field, and design dimensions such as holistic coverage of LLM architectures and models, use of public annotation guidelines, data imbalance, and the English language, have not been fully studied.

With respect to data imbalance, related work has served as groundwork for the methods used in Section~\ref{sec:data-imbalance}. Recent work shows that GPT models still underperform traditional machine learning on imbalanced security bug report identification~\citep{FRANCA2025107778}.
Recent studies show that loss-based methods can improve minority-class recognition in multi-label and domain-specific text classification, especially when label co-occurrence makes naive resampling less reliable~\citep{Ramakrishnan2025Improving,Huang2021Balancing}. Undersampling has likewise remained relevant as a distribution-level alternative, with evidence that carefully removing redundant majority examples can reduce classifier bias without harming effectiveness, although its benefit depends on which instances are discarded and excessive removal can eliminate useful information~\citep{Fonseca2024}. Oversampling, from SMOTE-style interpolation to text augmentation, can improve robustness under skewed label distributions, but outcomes remain dataset- and classifier-sensitive~\citep{Taskiran2025A,Henning2022A}.
Our study combines multiple data augmentation techniques and proposes an approach to rank and select a generative-based method to increase semantic coverage while preserving labelling scope, and evaluates a sample of the generated reviews in a human-as-a-judge setting to provide evidence on the reliability of such labels.

\section{Conclusions}\label{sec:conclusions}

This paper investigated automatic multi-label emotion classification of mobile app reviews using encoder-only and decoder-only large language models, together with a systematic comparison of imbalance-mitigation strategies.
The results show fine-grained emotion classification can be automated cautiously. Macro-F1 remains modest even for the best configuration tested: the multi-label encoder with generative augmentation and positive-weighted loss reaches $0.591$, versus $0.642$ for the best decoder, Gemini-3.5-Flash.
This varies by emotion. \emph{Joy}, \emph{Anticipation} and \emph{Sadness} are classified reliably under most configurations (F1 up to $0.83$--$0.84$ for the best decoder, Table~\ref{tab:rq2-per-emotion}). \emph{Surprise} stays hard everywhere tested (F1 $\leq 0.346$, Table~\ref{tab:rq4-per-emotion}), and \emph{Fear}, \emph{Disgust} and \emph{Anger}, though substantially improved by mitigation, still sit around F1 $0.41$--$0.50$. The remaining room for improvement concentrates on these harder emotions.
What makes automation worthwhile despite these limitations is cost. A mitigated fine-tuned encoder closes to within $0.051$ macro-F1 (multi-label) or $0.112$ (binary ensemble) of the best decoder at one to three orders of magnitude lower inference latency (milliseconds versus hundreds of milliseconds to over a second per sentence, Tables~\ref{tab:rq1-cost} and~\ref{tab:rq2-cost}).
This gap holds even for the binary-ensemble formulation's nine independently-trained classifiers, which together still run faster than a single decoder call.

An organisation using this configuration could set its own confidence threshold, once calibrated, automate high-confidence predictions, and route a sample of the rest to human review, depending on its risk tolerance and which emotions matter most downstream.
Relabelling that human-reviewed tail could produce targeted training data complementary to the augmentation evaluated here, closing a loop between automated classification and human judgement on ambiguous or semantically overlapping emotions, which even trained annotators do not fully agree on (Challenge~1)~\citep{Motger2025Emotions}.
Interpretations such as why \emph{Sadness} absorbs \emph{Anger}, \emph{Fear} and \emph{Surprise} under a shared multi-label head are tied to this corpus and taxonomy.
The methodological lessons may generalise further: a threshold-based label-assignment rule may make the comparison between model formulations sensitive to calibration, no single loss dominates across formulations, and generative augmentation outperforms loss reweighting and undersampling at this corpus scale.
We expect these properties to recur in other multi-label classification tasks under severe imbalance.

Several classification tasks across the software engineering lifecycle share this profile (many possible labels, most instances carrying one or two, and a handful of frequent labels dwarfing the rest), such as review-type classification, feature and requirement extraction from user feedback, bug-report severity and priority labelling, security-relevant issue classification, and self-admitted technical debt identification, wherever teams must choose, under a fixed budget, between fine-tuning a small encoder, prompting a large decoder, or investing in imbalance mitigation first. The comparative framework applied here, sweeping formulation, prompting and mitigation strategy jointly and reading results through both effectiveness and cost, offers a template for that decision beyond app-review emotion classification.

\section*{Data Availability Statement}

The code, data and trained models associated with this study are publicly available.
The full experimental pipeline, configuration files, and ground-truth and synthetic
corpora are released as a replication package on \href{https://github.com/nlp4se/EmotionClassification}{\simpleicon{github}\ GitHub}.
The two best-performing encoder-only checkpoints identified in this study (i.e., the
multi-label and binary-ensemble formulations) are released as a curated \href{https://huggingface.co/collections/quim-motger/emotion-classification-from-mobile-app-reviews-6aa28a181753ea666e10a8c7}{\simpleicon{huggingface}\ Hugging Face
Hub collection}.

\printcredits

\section*{Declaration of generative AI and AI-assisted technologies in the manuscript preparation process}

During the preparation of this work the authors used generative AI tools for language editing, limited code assistance and replication-package documentation. The authors reviewed and edited the content as needed and take full responsibility for the published article.


\bibliographystyle{cas-model2-names}
\bibliography{cas-refs}

@inproceedings{Motger2025Emotions,
  author = {Motger, Quim and Oriol, Marc and Tiessler, Max and Franch, Xavier and Marco, Jordi},
  title = {What About Emotions? {G}uiding Fine-Grained Emotion Extraction from Mobile App Reviews},
  booktitle = {2025 IEEE 33rd International Requirements Engineering Conference (RE)},
  year = {2025},
  pages = {6--18},
  doi = {10.1109/RE63999.2025.00012},
  publisher = {IEEE},
}

@inproceedings{Motger2024TFREX,
  author = {Motger, Quim and Miaschi, Alessio and Dell'Orletta, Felice and Franch, Xavier and Marco, Jordi},
  title = {{T-FREX}: A Transformer-based Feature Extraction Method from Mobile App Reviews},
  booktitle = {2024 IEEE International Conference on Software Analysis, Evolution and Reengineering (SANER)},
  year = {2024},
  pages = {227--238},
  doi = {10.1109/SANER60148.2024.00030},
}

@article{Motger2025TFREXEMSE,
  author = {Motger, Quim and Miaschi, Alessio and Dell'Orletta, Felice and Franch, Xavier and Marco, Jordi},
  title = {Leveraging Encoder-only Large Language Models for Mobile App Review Feature Extraction},
  journal = {Empirical Software Engineering},
  year = {2025},
  volume = {30},
  number = {4},
  pages = {104},
  doi = {10.1007/s10664-025-10660-y},
  publisher = {Springer},
}

@article{Dabrowski2022,
  author = {D{\k{a}}browski, Jacek and Letier, Emmanuel and Perini, Anna and Susi, Angelo},
  title = {Analysing App Reviews for Software Engineering: A Systematic Literature Review},
  journal = {Empirical Software Engineering},
  volume = {27},
  number = {2},
  pages = {43},
  year = {2022},
  doi = {10.1007/s10664-021-10065-7},
}

@inproceedings{Maalej2015,
  author = {Maalej, Walid and Nabil, Hadeer},
  title = {Bug Report, Feature Request, or Simply Praise? {O}n Automatically Classifying App Reviews},
  booktitle = {2015 IEEE 23rd International Requirements Engineering Conference (RE)},
  year = {2015},
  pages = {116--125},
  doi = {10.1109/RE.2015.7320414},
}

@inproceedings{Guzman2014,
  author = {Guzman, Emitza and Maalej, Walid},
  title = {How Do Users Like This Feature? {A} Fine Grained Sentiment Analysis of App Reviews},
  booktitle = {2014 IEEE 22nd International Requirements Engineering Conference (RE)},
  year = {2014},
  pages = {153--162},
  doi = {10.1109/RE.2014.6912257},
}

@inproceedings{Pagano2013,
  author = {Pagano, Dennis and Maalej, Walid},
  title = {User Feedback in the {AppStore}: An Empirical Study},
  booktitle = {2013 21st IEEE International Requirements Engineering Conference (RE)},
  year = {2013},
  pages = {125--134},
  doi = {10.1109/RE.2013.6636712},
}

@article{Lin2022,
  author = {Lin, Bin and Cassee, Nathan and Serebrenik, Alexander and Bavota, Gabriele and Novielli, Nicole and Lanza, Michele},
  title = {Opinion Mining for Software Development: A Systematic Literature Review},
  journal = {ACM Transactions on Software Engineering and Methodology},
  volume = {31},
  number = {3},
  pages = {38:1--38:41},
  year = {2022},
  doi = {10.1145/3490388},
  publisher = {ACM},
}

@article{Araujo2022,
  author = {Araujo, Adailton F. and G{\^o}lo, Marcos P. S. and Marcacini, Ricardo M.},
  title = {Opinion Mining for App Reviews: An Analysis of Textual Representation and Predictive Models},
  journal = {Automated Software Engineering},
  volume = {29},
  number = {1},
  pages = {5},
  year = {2022},
  doi = {10.1007/s10515-021-00301-1},
  publisher = {Springer},
}

@article{Alturayeif2023,
  author = {Alturayeif, Nouf and Aljamaan, Hamoud and Hassine, Jameleddine},
  title = {An Automated Approach to Aspect-Based Sentiment Analysis of Apps Reviews Using Machine and Deep Learning},
  journal = {Automated Software Engineering},
  volume = {30},
  number = {2},
  pages = {30},
  year = {2023},
  doi = {10.1007/s10515-023-00397-7},
  publisher = {Springer},
}

@inproceedings{Alturaief2021,
  author = {Alturaief, Nouf and Aljamaan, Hamoud and Baslyman, Malak},
  title = {{AWARE}: Aspect-Based Sentiment Analysis Dataset of Apps Reviews for Requirements Elicitation},
  booktitle = {2021 36th IEEE/ACM International Conference on Automated Software Engineering Workshops (ASEW)},
  year = {2021},
  pages = {211--218},
  doi = {10.1109/ASEW52652.2021.00049},
}

@inproceedings{Johann2017,
  author = {Johann, Timo and Stanik, Christoph and {Alizadeh B.}, Alireza M. and Maalej, Walid},
  title = {{SAFE}: A Simple Approach for Feature Extraction from App Descriptions and App Reviews},
  booktitle = {2017 IEEE 25th International Requirements Engineering Conference (RE)},
  year = {2017},
  pages = {21--30},
  doi = {10.1109/RE.2017.71},
}

@article{Plutchik2001,
  author = {Plutchik, Robert},
  title = {The Nature of Emotions},
  journal = {American Scientist},
  volume = {89},
  number = {4},
  pages = {344--350},
  year = {2001},
}

@article{Riccosan2023,
  author = {Riccosan and Saputra, Karen Etania},
  title = {Multilabel Multiclass Sentiment and Emotion Dataset from {Indonesian} Mobile Application Review},
  journal = {Data in Brief},
  volume = {50},
  pages = {109576},
  year = {2023},
  doi = {10.1016/j.dib.2023.109576},
}

@inproceedings{Siqueira2024,
  author = {Siqueira, Vitor X. and Costa, Ricardo Luiz Hentges and Soares, Tales Schifelbein and Lunardi, Gabriel M. and Silva, Williamson},
  title = {Dataset Anotado de Sentimentos a partir de coment{\'a}rios de Aplicativos M{\'o}veis},
  booktitle = {Anais do {VI} Dataset Showcase Workshop ({DSW} 2024)},
  year = {2024},
  pages = {65--76},
  publisher = {Sociedade Brasileira de Computa{\c{c}}{\~a}o},
  doi = {10.5753/dsw.2024.243926},
}

@article{Singh2022,
  author = {Singh, Yashdeep and Suri, Pradeep Kumar},
  title = {An Empirical Analysis of Mobile Learning App Usage Experience},
  journal = {Technology in Society},
  volume = {68},
  pages = {101929},
  year = {2022},
  doi = {10.1016/j.techsoc.2022.101929},
  publisher = {Elsevier},
}

@inproceedings{Keertipati2016,
  author = {Keertipati, Swetha and Savarimuthu, Bastin Tony Roy and Licorish, Sherlock A.},
  title = {Approaches for Prioritizing Feature Improvements Extracted from App Reviews},
  booktitle = {Proceedings of the 20th International Conference on Evaluation and Assessment in Software Engineering ({EASE} '16)},
  year = {2016},
  pages = {33:1--33:6},
  publisher = {ACM},
  doi = {10.1145/2915970.2916003},
}

@misc{Malgaonkar2019,
  author = {Malgaonkar, Saurabh and Lee, C. W. and Licorish, Sherlock A. and Savarimuthu, Bastin Tony Roy and Tahir, Amjed},
  title = {{AppSent}: A Tool That Analyzes App Reviews},
  howpublished = {arXiv preprint arXiv:1907.10191},
  year = {2019},
}

@article{Savarimuthu2023,
  author = {Savarimuthu, Bastin Tony Roy and Corbett, Jacqueline and Yasir, Muhammad and Lakshmi, Vijaya},
  title = {Improving Information Systems Sustainability by Applying Machine Learning to Detect and Reduce Data Waste},
  journal = {Communications of the Association for Information Systems},
  volume = {53},
  pages = {189--213},
  year = {2023},
  doi = {10.17705/1CAIS.05308},
}

@article{Cabellos2022,
  author = {Cabellos, Beatriz and Pozo, Juan-Ignacio and Mar{\'i}n-Rubio, Kevin and S{\'a}nchez, Daniel L.},
  title = {Do Pro-Social Video Games Promote Moral Activity?: An Analysis of User Reviews of {Papers, Please}},
  journal = {Education and Information Technologies},
  volume = {27},
  number = {8},
  pages = {11411--11442},
  year = {2022},
  doi = {10.1007/s10639-022-11072-x},
}

@article{Hidellaarachchi2023,
  author = {Hidellaarachchi, Dulaji and Grundy, John and Hoda, Rashina and Mueller, Ingo},
  title = {The Influence of Human Aspects on Requirements Engineering-Related Activities: Software Practitioners' Perspective},
  journal = {ACM Transactions on Software Engineering and Methodology},
  volume = {32},
  number = {5},
  pages = {108:1--108:37},
  year = {2023},
  doi = {10.1145/3546943},
  publisher = {ACM},
}

@inproceedings{Cheng2023,
  author = {Cheng, Ben and Arora, Chetan and Liu, Xiao and Hoang, Thuong and Wang, Yi and Grundy, John},
  title = {Multi-Modal Emotion Recognition for Enhanced Requirements Engineering: A Novel Approach},
  booktitle = {2023 IEEE 31st International Requirements Engineering Conference (RE)},
  year = {2023},
  pages = {299--304},
  doi = {10.1109/RE57278.2023.00039},
}

@article{Hou2024,
  author = {Hou, Xinyi and Zhao, Yanjie and Liu, Yue and Yang, Zhou and Wang, Kailong and Li, Li and Luo, Xiapu and Lo, David and Grundy, John and Wang, Haoyu},
  title = {Large Language Models for Software Engineering: A Systematic Literature Review},
  journal = {ACM Transactions on Software Engineering and Methodology},
  volume = {33},
  number = {8},
  pages = {220:1--220:79},
  year = {2024},
  doi = {10.1145/3695988},
}

@article{Zhao2023,
  author = {Zhao, Wayne Xin and Zhou, Kun and Li, Junyi and Tang, Tianyi and Dong, Zican and Hou, Yupeng and Zhang, Beichen and Min, Yingqian and Zhang, Junjie and Liu, Peiyu and Wang, Xiaolei and Du, Yifan and Yang, Chen and Chen, Yushuo and Chen, Zhipeng and Jiang, Jinhao and Ren, Ruiyang and Li, Yifan and Tang, Xinyu and Liu, Zikang and Hu, Yiwen and Nie, Jian-Yun and Wen, Ji-Rong},
  title = {A Survey of Large Language Models},
  journal = {Frontiers of Computer Science},
  volume = {20},
  number = {12},
  pages = {2012627},
  year = {2026},
  doi = {10.1007/s11704-026-60308-3},
}

@article{Zhang2025Sentiment,
  author = {Zhang, Ting and Irsan, Ivana Clairine and Thung, Ferdian and Lo, David},
  title = {Revisiting Sentiment Analysis for Software Engineering in the Era of Large Language Models},
  journal = {ACM Transactions on Software Engineering and Methodology},
  volume = {34},
  number = {3},
  pages = {60:1--60:30},
  year = {2025},
  doi = {10.1145/3697009},
  publisher = {ACM},
}

@inproceedings{Brown2020,
  author = {Brown, Tom and Mann, Benjamin and Ryder, Nick and Subbiah, Melanie and Kaplan, Jared D and Dhariwal, Prafulla and Neelakantan, Arvind and Shyam, Pranav and Sastry, Girish and Askell, Amanda and others},
  title = {Language Models are Few-Shot Learners},
  booktitle = {Advances in Neural Information Processing Systems},
  volume = {33},
  pages = {1877--1901},
  year = {2020},
}

@inproceedings{Devlin2019,
  author = {Devlin, Jacob and Chang, Ming-Wei and Lee, Kenton and Toutanova, Kristina},
  title = {{BERT}: Pre-training of Deep Bidirectional Transformers for Language Understanding},
  booktitle = {Proceedings of the 2019 Conference of the North {A}merican Chapter of the Association for Computational Linguistics: Human Language Technologies},
  year = {2019},
  pages = {4171--4186},
  publisher = {Association for Computational Linguistics},
  doi = {10.18653/v1/N19-1423},
}

@misc{Liu2019RoBERTa,
  author = {Liu, Yinhan and Ott, Myle and Goyal, Naman and Du, Jingfei and Joshi, Mandar and Chen, Danqi and Levy, Omer and Lewis, Mike and Zettlemoyer, Luke and Stoyanov, Veselin},
  title = {{RoBERTa}: A Robustly Optimized {BERT} Pretraining Approach},
  howpublished = {arXiv preprint arXiv:1907.11692},
  year = {2019},
}

@inproceedings{Yang2019XLNet,
  author = {Yang, Zhilin and Dai, Zihang and Yang, Yiming and Carbonell, Jaime and Salakhutdinov, Ruslan and Le, Quoc V.},
  title = {{XLNet}: Generalized Autoregressive Pretraining for Language Understanding},
  booktitle = {Advances in Neural Information Processing Systems},
  volume = {32},
  year = {2019},
}

@inproceedings{Sanh2019DistilBERT,
  author = {Sanh, Victor and Debut, Lysandre and Chaumond, Julien and Wolf, Thomas},
  title = {{DistilBERT}, a Distilled Version of {BERT}: Smaller, Faster, Cheaper and Lighter},
  booktitle = {5th Workshop on Energy Efficient Machine Learning and Cognitive Computing ({EMC}$^2$) at {NeurIPS} 2019},
  year = {2019},
  url = {https://www.emc2-ai.org/assets/docs/neurips-19/emc2-neurips19-paper-33.pdf},
}

@inproceedings{He2023DeBERTaV3,
  author = {He, Pengcheng and Gao, Jianfeng and Chen, Weizhu},
  title = {{DeBERTaV3}: Improving {DeBERTa} using {ELECTRA}-Style Pre-Training with Gradient-Disentangled Embedding Sharing},
  booktitle = {International Conference on Learning Representations (ICLR)},
  year = {2023},
}

@inproceedings{Vaswani2017,
  author = {Vaswani, Ashish and Shazeer, Noam and Parmar, Niki and Uszkoreit, Jakob and Jones, Llion and Gomez, Aidan N. and Kaiser, {\L}ukasz and Polosukhin, Illia},
  title = {Attention Is All You Need},
  booktitle = {Advances in Neural Information Processing Systems},
  volume = {30},
  year = {2017},
}

@inproceedings{Lin2017Focal,
  author = {Lin, Tsung-Yi and Goyal, Priya and Girshick, Ross and He, Kaiming and Doll{\'a}r, Piotr},
  title = {Focal Loss for Dense Object Detection},
  booktitle = {Proceedings of the IEEE International Conference on Computer Vision (ICCV)},
  year = {2017},
  pages = {2980--2988},
  doi = {10.1109/ICCV.2017.324},
}

@article{He2009Imbalance,
  author = {He, Haibo and Garcia, Edwardo A.},
  title = {Learning from Imbalanced Data},
  journal = {IEEE Transactions on Knowledge and Data Engineering},
  volume = {21},
  number = {9},
  pages = {1263--1284},
  year = {2009},
  doi = {10.1109/TKDE.2008.239},
  publisher = {IEEE},
}

@inproceedings{Sechidis2011,
  author = {Sechidis, Konstantinos and Tsoumakas, Grigorios and Vlahavas, Ioannis},
  title = {On the Stratification of Multi-Label Data},
  booktitle = {Machine Learning and Knowledge Discovery in Databases (ECML PKDD 2011)},
  series = {Lecture Notes in Computer Science},
  volume = {6913},
  pages = {145--158},
  year = {2011},
  publisher = {Springer},
  doi = {10.1007/978-3-642-23808-6_10},
}

@article{Charte2015MLSMOTE,
  author = {Charte, Francisco and Rivera, Antonio J. and del Jesus, Mar{\'i}a J. and Herrera, Francisco},
  title = {{MLSMOTE}: Approaching Imbalanced Multilabel Learning through Synthetic Instance Generation},
  journal = {Knowledge-Based Systems},
  volume = {89},
  pages = {385--397},
  year = {2015},
  doi = {10.1016/j.knosys.2015.07.024},
}

@inproceedings{SemEval2025Focal,
  author = {Kesanam, Ashinee and Ravi Ram, Gummuluri Venkata and Banoth, Chaithanya Swaroop and Reddy, G. Rama Mohana},
  title = {{NITK-VITAL} at {SemEval}-2025 Task 11: {Focal-RoBERTa}: Addressing Class Imbalance in Multi-Label Emotion Classification},
  booktitle = {Proceedings of the 19th International Workshop on Semantic Evaluation (SemEval-2025)},
  year = {2025},
  pages = {1077--1081},
  address = {Vienna, Austria},
  publisher = {Association for Computational Linguistics},
  url = {https://aclanthology.org/2025.semeval-1.142/},
}

@article{Labd2025AFL,
  author = {Labd, Zakia and Bahassine, Said and Housni, Khalid},
  title = {{AFL-BERT}: Enhancing Minority Class Detection in Multi-Label Text Classification with Adaptive Focal Loss and {BERT}},
  journal = {International Journal of Advanced Computer Science and Applications},
  volume = {16},
  number = {7},
  year = {2025},
  doi = {10.14569/IJACSA.2025.0160749},
}

@inproceedings{Cui2025Weighted,
  author = {Cui, Xia},
  title = {xiacui at {SemEval}-2025 Task 11: Addressing Data Imbalance in Transformer-Based Multi-Label Emotion Detection with Weighted Loss},
  booktitle = {Proceedings of the 19th International Workshop on Semantic Evaluation (SemEval-2025)},
  year = {2025},
  pages = {247--256},
  address = {Vienna, Austria},
  publisher = {Association for Computational Linguistics},
  url = {https://aclanthology.org/2025.semeval-1.37/},
}

@inproceedings{Reimers2019,
  author = {Reimers, Nils and Gurevych, Iryna},
  title = {Sentence-{BERT}: Sentence Embeddings using {Siamese} {BERT}-Networks},
  booktitle = {Proceedings of the 2019 Conference on Empirical Methods in Natural Language Processing and the 9th International Joint Conference on Natural Language Processing (EMNLP-IJCNLP)},
  year = {2019},
  pages = {3982--3992},
  publisher = {Association for Computational Linguistics},
  doi = {10.18653/v1/D19-1410},
}

@inproceedings{Wang2020MiniLM,
  author = {Wang, Wenhui and Wei, Furu and Dong, Li and Bao, Hangbo and Yang, Nan and Zhou, Ming},
  title = {{MiniLM}: Deep Self-Attention Distillation for Task-Agnostic Compression of Pre-Trained Transformers},
  booktitle = {Advances in Neural Information Processing Systems},
  volume = {33},
  pages = {5776--5788},
  year = {2020},
}

@book{Wohlin2012,
  author = {Wohlin, Claes and Runeson, Per and H{\"o}st, Martin and Ohlsson, Magnus C. and Regnell, Bj{\"o}rn and Wessl{\'e}n, Anders},
  title = {Experimentation in Software Engineering},
  publisher = {Springer},
  address = {Berlin, Heidelberg},
  year = {2012},
  doi = {10.1007/978-3-642-29044-2},
}

@inproceedings{Shaw2003,
  author = {Shaw, Mary},
  title = {Writing Good Software Engineering Research Papers},
  booktitle = {Proceedings of the 25th International Conference on Software Engineering (ICSE)},
  year = {2003},
  pages = {726--736},
  doi = {10.1109/ICSE.2003.1201262},
}

@inproceedings{Loshchilov2019,
  author = {Loshchilov, Ilya and Hutter, Frank},
  title = {Decoupled Weight Decay Regularization},
  booktitle = {International Conference on Learning Representations (ICLR)},
  year = {2019},
}

@article{Tsoumakas2007,
  author = {Tsoumakas, Grigorios and Katakis, Ioannis},
  title = {Multi-Label Classification: An Overview},
  journal = {International Journal of Data Warehousing and Mining},
  volume = {3},
  number = {3},
  pages = {1--13},
  year = {2007},
  doi = {10.4018/jdwm.2007070101},
}

@inproceedings{AnabyTavor2020,
  author = {Anaby-Tavor, Ateret and Carmeli, Boaz and Goldbraich, Esther and Kantor, Amir and Kour, George and Shlomov, Segev and Tepper, Naama and Zwerdling, Naama},
  title = {Do Not Have Enough Data? {D}eep Learning to the Rescue!},
  booktitle = {Proceedings of the AAAI Conference on Artificial Intelligence},
  volume = {34},
  pages = {7383--7390},
  year = {2020},
  doi = {10.1609/aaai.v34i05.6233},
}

@misc{statcounter2025chatbots,
  author = {{StatCounter}},
  title = {{AI} Chatbot Market Share Worldwide},
  howpublished = {\url{https://gs.statcounter.com/ai-chatbot-market-share}},
  year = {2026},
  note = {Accessed 2026-07-24},
}

@misc{artificialanalysis2026,
  author = {{Artificial Analysis}},
  title = {Artificial Analysis Intelligence Index},
  howpublished = {\url{https://artificialanalysis.ai/}},
  year = {2026},
  note = {Accessed 2026-07-24},
}

@article{AMEER2023118534,
  author = {Ameer, Iqra and B{\"o}l{\"u}c{\"u}, Necva and Siddiqui, Muhammad Hammad Fahim and Can, Burcu and Sidorov, Grigori and Gelbukh, Alexander},
  title = {Multi-label Emotion Classification in Texts Using Transfer Learning},
  journal = {Expert Systems with Applications},
  volume = {213},
  pages = {118534},
  year = {2023},
  doi = {10.1016/j.eswa.2022.118534},
}

@article{Lalk2025,
  author = {Lalk, Christopher and Targan, Kim and Steinbrenner, Tobias and Schaffrath, Jana and Eberhardt, Steffen and Schwartz, Brian and Vehlen, Antonia and Lutz, Wolfgang and Rubel, Julian},
  title = {Employing Large Language Models for Emotion Detection in Psychotherapy Transcripts},
  journal = {Frontiers in Psychiatry},
  volume = {16},
  pages = {1504306},
  year = {2025},
  doi = {10.3389/fpsyt.2025.1504306},
}

@inproceedings{reichman2025emotions,
  author = {Reichman, Benjamin and Avsian, Adar and Heck, Larry},
  title = {Emotions Where Art Thou: Understanding and Characterizing the Emotional Latent Space of Large Language Models},
  booktitle = {International Conference on Learning Representations (ICLR)},
  year = {2026},
}

@inproceedings{Imran2022,
  author = {Imran, Mia Mohammad and Jain, Yashasvi and Chatterjee, Preetha and Damevski, Kostadin},
  title = {Data Augmentation for Improving Emotion Recognition in Software Engineering Communication},
  booktitle = {Proceedings of the 37th IEEE/ACM International Conference on Automated Software Engineering (ASE '22)},
  year = {2023},
  pages = {29:1--29:13},
  publisher = {ACM},
  doi = {10.1145/3551349.3556925},
}

@article{FRANCA2025107778,
  author = {Fran{\c{c}}a, Hor{\'a}cio L. and Goseva-Popstojanova, Katerina and Teixeira, C{\'e}sar and Laranjeiro, Nuno},
  title = {{GPTs} Are Not the Silver Bullet: Performance and Challenges of Using {GPTs} for Security Bug Report Identification},
  journal = {Information and Software Technology},
  volume = {185},
  pages = {107778},
  year = {2025},
  doi = {10.1016/j.infsof.2025.107778},
}

@inproceedings{Fonseca2024,
  author = {Fonseca, Guilherme and Prenassi, Gabriel and Cunha, Washington and Gon{\c{c}}alves, Marcos Andr{\'e} and Rocha, Leonardo},
  title = {Estrat{\'e}gias de Undersampling para Redu{\c{c}}{\~a}o de Vi{\'e}s em Classifica{\c{c}}{\~a}o de Texto Baseada em Transformers},
  booktitle = {Proceedings of the 30th Brazilian Symposium on Multimedia and the Web (WebMedia 2024)},
  year = {2024},
  pages = {144--152},
  publisher = {Sociedade Brasileira de Computa{\c{c}}{\~a}o},
  doi = {10.5753/webmedia.2024.241229},
}

@inproceedings{Henning2022A,
  author = {Henning, Sophie and Beluch, William and Fraser, Alexander and Friedrich, Annemarie},
  title = {A Survey of Methods for Addressing Class Imbalance in Deep-Learning Based Natural Language Processing},
  booktitle = {Proceedings of the 17th Conference of the European Chapter of the Association for Computational Linguistics (EACL)},
  year = {2023},
  pages = {523--540},
  publisher = {Association for Computational Linguistics},
  doi = {10.18653/v1/2023.eacl-main.38},
}

@inproceedings{Huang2021Balancing,
  author = {Huang, Yi and Giledereli, Buse and K{\"o}ksal, Abdullatif and {\"O}zg{\"u}r, Arzucan and Ozkirimli, Elif},
  title = {Balancing Methods for Multi-label Text Classification with Long-Tailed Class Distribution},
  booktitle = {Proceedings of the 2021 Conference on Empirical Methods in Natural Language Processing (EMNLP)},
  year = {2021},
  pages = {8153--8161},
  publisher = {Association for Computational Linguistics},
  doi = {10.18653/v1/2021.emnlp-main.643},
}

@article{Ramakrishnan2025Improving,
  author = {Ramakrishnan, Sandhya and Dhinesh Babu, L. D.},
  title = {Improving Multi-Label Emotion Classification on Imbalanced Social Media Data With {BERT} and Clipped Asymmetric Loss},
  journal = {IEEE Access},
  year = {2025},
  volume = {13},
  pages = {60589--60601},
  doi = {10.1109/ACCESS.2025.3557091},
}

@article{Taskiran2025A,
  author = {Taskiran, Salimkan Fatma and Turkoglu, Bahaeddin and Kaya, Ersin and A{\c{s}}uro{\u{g}}lu, Tun{\c{c}}},
  title = {A Comprehensive Evaluation of Oversampling Techniques for Enhancing Text Classification Performance},
  journal = {Scientific Reports},
  year = {2025},
  volume = {15},
  pages = {21631},
  doi = {10.1038/s41598-025-05791-7},
}

@inproceedings{Ye2022ZeroGen,
  author = {Ye, Jiacheng and Gao, Jiahui and Li, Qintong and Xu, Hang and Feng, Jiangtao and Wu, Zhiyong and Yu, Tao and Kong, Lingpeng},
  title = {{ZeroGen}: Efficient Zero-shot Learning via Dataset Generation},
  booktitle = {Proceedings of the 2022 Conference on Empirical Methods in Natural Language Processing (EMNLP)},
  year = {2022},
  pages = {11653--11669},
  publisher = {Association for Computational Linguistics},
  doi = {10.18653/v1/2022.emnlp-main.801},
}

@inproceedings{Hsieh2023Distilling,
  author = {Hsieh, Cheng-Yu and Li, Chun-Liang and Yeh, Chih-kuan and Nakhost, Hootan and Fujii, Yasuhisa and Ratner, Alex and Krishna, Ranjay and Lee, Chen-Yu and Pfister, Tomas},
  title = {Distilling Step-by-Step! {O}utperforming Larger Language Models with Less Training Data and Smaller Model Sizes},
  booktitle = {Findings of the Association for Computational Linguistics: ACL 2023},
  year = {2023},
  pages = {8003--8017},
  publisher = {Association for Computational Linguistics},
  doi = {10.18653/v1/2023.findings-acl.507},
}

@inproceedings{Yu2023AttrPrompt,
  author = {Yu, Yue and Zhuang, Yuchen and Zhang, Jieyu and Meng, Yu and Ratner, Alexander J. and Krishna, Ranjay and Shen, Jiaming and Zhang, Chao},
  title = {Large Language Model as Attributed Training Data Generator: A Tale of Diversity and Bias},
  booktitle = {Advances in Neural Information Processing Systems 36 (NeurIPS 2023), Datasets and Benchmarks Track},
  year = {2023},
  pages = {55734--55784},
}

\end{document}